\documentclass[11pt]{article}

\usepackage[preprint]{acl}

\usepackage{times}
\usepackage{latexsym}

\usepackage[T1]{fontenc}

\usepackage[utf8]{inputenc}

\usepackage{microtype}

\usepackage{inconsolata}

\usepackage{graphicx}
\usepackage{CJKutf8}
\usepackage{booktabs}
\usepackage{multirow}
\usepackage[table]{xcolor}
\definecolor{Gray}{gray}{0.92}
\usepackage{makecell}
\usepackage{amsmath}
\usepackage{amsfonts}
\usepackage[table]{xcolor}
\usepackage{graphicx}
\usepackage{float}
\usepackage{cuted}
\title{Incentivizing Parametric Knowledge via Reinforcement Learning with Verifiable Rewards for Cross-Cultural Entity Translation}

\author{
 \textbf{Jiang Zhou\textsuperscript{1}},
 \textbf{Xiaohu Zhao\textsuperscript{2}},
 \textbf{Xinwei Wu\textsuperscript{1}},
 \textbf{Tianyu Dong\textsuperscript{1}},
\\
 \textbf{Hao Wang\textsuperscript{2}},
 \textbf{Yangyang Liu\textsuperscript{2}},
 \textbf{Heng Liu\textsuperscript{2}},
 \textbf{Linlong Xu\textsuperscript{2}},
\\
 \textbf{Longyue Wang\textsuperscript{2}},
 \textbf{Weihua Luo\textsuperscript{2}},
 \textbf{Deyi Xiong\textsuperscript{1}$^\dagger$}
\\
\\
 \textsuperscript{1}TJUNLP Lab, Tianjin University, China
 \\
 \textsuperscript{2}Alibaba Group, China
\\
 \small{
    \href{mailto:dyxiong@tju.edu.cn}{dyxiong@tju.edu.cn}
 }
}

\begin{document}
\maketitle

\let\oldthefootnote\thefootnote

\let\thefootnote\relax\footnotetext{$^\dagger$ Corresponding Author.}

\let\thefootnote\relax\footnotetext{}
\let\thefootnote\oldthefootnote

\begin{abstract}
Cross-cultural entity translation remains challenging for large language models (LLMs) as literal or phonetic renderings are usually yielded instead of culturally appropriate translations in context. However, relevant knowledge may already be encoded in model parameters during large-scale pre-training. To incentivize the effective use of parametric knowledge, we propose \textbf{EA-RLVR} (Entity-Anchored Reinforcement Learning with Verifiable Rewards), a training framework that optimizes cross-cultural entity translation without relying on external knowledge bases. EA-RLVR anchors supervision on a verifiable, entity-level reward signal and incorporates lightweight structural gates to stabilize optimization. This design steers the model toward learning a robust reasoning process rather than merely imitating reference translations.
We evaluate EA-RLVR on XC-Translate and observe consistent improvements in both entity translation accuracy and out-of-domain generalization. Specifically, training on merely 7k samples boosts Qwen3-14B’s entity translation accuracy from 23.66\% to 31.87\% on a 50k test set comprising \textbf{entirely unseen entities}.
The learned entity translation ability also transfers to general translation, yielding +1.35 XCOMET on WMT24pp, which scales to +1.59 with extended optimization.
Extensive analyses of $pass@k$ dynamics and reward formulations attribute these gains to superior sampling efficiency and a stable optimization landscape. 
\end{abstract}

\section{Introduction}
\begin{figure}
    \centering
    \includegraphics[width=1\linewidth]{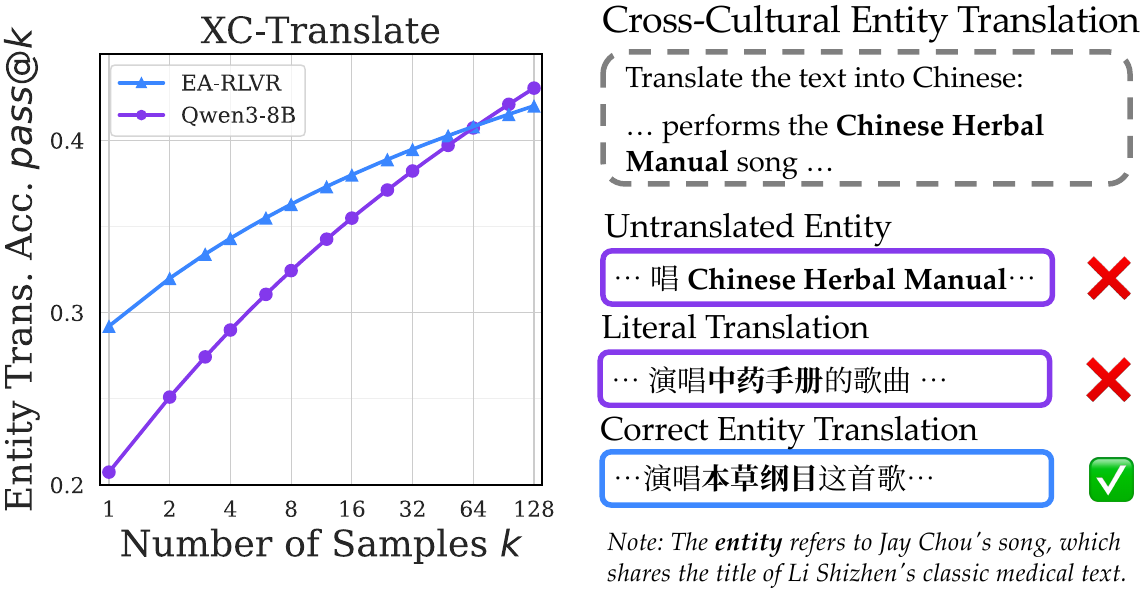}
    \caption{\textbf{(Left)} Entity translation accuracy (\%) $pass@k$ curves demonstrate the base model possesses latent knowledge (high accuracy at large $k$) that EA-RLVR effectively activates at $k=1$ . 
    \textbf{(Right)} An illustration of the challenge in cross-cultural entity translation.}
    \label{fig:teaser}
\end{figure}
At its core, machine translation aspires to make culturally situated texts accessible across languages. Despite substantial progress with multilingual large language models \citep{pan2025advancing}, current systems often fall short of this goal in settings where translation hinges on culturally grounded entities such as books, films, places, songs and idioms~\citep{yao-etal-2024-benchmarking}. In these cases, producing an accurate, culture-aligned translation requires identifying, in context, which real-world entity is being referred to and how it is conventionally named in the target culture~\citep{moghe-etal-2025-machine}. Recent evaluations have shown that even frontier proprietary LLMs frequently default to literal or phonetic renderings that are grammatically well formed but semantically inappropriate in context, thereby altering or obscuring the intended meaning of the source text~\citep{conia2024xctranslate}.

A widely adopted workaround for this limitation is to equip translation systems with external knowledge, e.g., through online retrieval, knowledge graphs, or curated databases~\citep{conia2024xctranslate,khandelwalnearest}. These approaches can improve accuracy when relevant information is successfully retrieved. However, they also introduce practical and structural constraints. The performance of such systems depends critically on how well the task aligns with underlying database~\citep{agrawal-etal-2023-context}, and in practice often requires task-specific retrievers that must be trained or tuned~\citep{wang-etal-2025-retrieval}. Moreover, it fundamentally shifts the bottleneck from contextual entity reasoning to the structure and coverage of the external knowledge source, making translation quality contingent on what can be retrieved.

On the other hand, as trained on corpora spanning trillions of tokens across diverse domains and languages, LLMs implicitly encode a wide range of entity correspondences, cultural references, and real-world usage conventions~\citep{2025qwen3,qwen25}. In principle, such knowledge should support cross-cultural translation. As illustrated in Figure~\ref{fig:teaser} (Left), the correct cultural entities are often present in the base model's probability distribution, evidenced by high accuracy when multiple sampling attempts ($pass@128$). However, such knowledge remains effectively inaccessible during standard single-pass generation ($pass@1$). Consequently, models frequently default to verbatim copying or literal renderings that obscure the intended meaning, such as retaining the source term or translating a song title literally as a medical manual (Figure~\ref{fig:teaser}, Right). These observations suggest that the core difficulty lies less in the availability of knowledge itself, but more in the absence of mechanisms that incentivize the model to surface that knowledge in a context-sensitive manner. 

To incentivize LLMs to leverage their parametric knowledge effectively, we propose \textbf{EA-RLVR} (\textbf{E}ntity-\textbf{A}nchored RL with \textbf{V}erifiable \textbf{R}ewards), 
a framework for cross-cultural entity translation driven by fully rule-based, automatically verifiable reward.  We cast cross-cultural entity translation as a sequence decision problem: given a source sentence, the model produces its own candidate translations, and a deterministic verifier evaluates whether the output expresses the correct target-culture entities. Rather than imitating reference translations, the model learns from verifiable rewards assigned to its own trajectories, reinforcing the reasoning that produces the correct entities. Concretely, EA-RLVR uses an \textbf{entity-matching reward} based on normalized substring matching between the predicted entity and the gold entity set. To stabilize optimization and reduce degenerate behaviors, we further introduce \textbf{structural gates} that modulate the reward according to lightweight output constraints (e.g., a prescribed reasoning format and translation length). This design avoids neural reward models that require additional computation and can be vulnerable to reward hacking in long-horizon RL, and it also addresses our empirical finding that neural metrics fails to provide supervision for culturally grounded entity choices. With these verifiable rewards and an efficient critic-free policy optimization recipe, EA-RLVR established a stable RLVR training framework for cross-cultural entity translation.

We conduct extensive experiments to evaluate EA-RLVR, yielding three key insights into its efficacy and underlying mechanisms:
\textbf{(1) EA-RLVR incentivizes parametric knowledge.}  
Training on only 7k examples generalizes to a 50k test set whose entities are entirely unseen during training, improving entity translation accuracy by +8.21\%--9.06\% across different model scales. 
\textbf{(2) The learned strategy transfers beyond entity evaluation.}  
On WMT24pp, our models achieve improvements of XCOMET by +1.25–1.35 points, even though XCOMET is never used as supervision. When scaling training to the full dataset and extending optimization to 1,000 steps, the gains increase to +1.59–1.68 points.
\textbf{(3) In-depth analyses clarify the dynamics of learning.}  
$pass@k$ evaluation, neural reward comparison, cross-lingual generalization, and examinations of reward-hacking behavior all point to the same pattern: our method improves sampling efficiency and induces stable, cross-cultural translation strategies, rather than encouraging memorization.

Our contributions are as follows: \textbf{(1)} We propose a novel framework for cross-cultural machine translation based on RLVR, showing that entity translation can be improved without access to external databases by directly incentivizing context-appropriate entity choices. \textbf{(2)} We empirically demonstrate that this approach improves entity translation accuracy and general translation quality across languages and model scales, including settings involving entirely unseen entities. \textbf{(3)} We provide in-depth analyses that explain the mechanism behind these improvements.

\begin{figure*}[th!]
    \centering
    \includegraphics[width=1\linewidth]{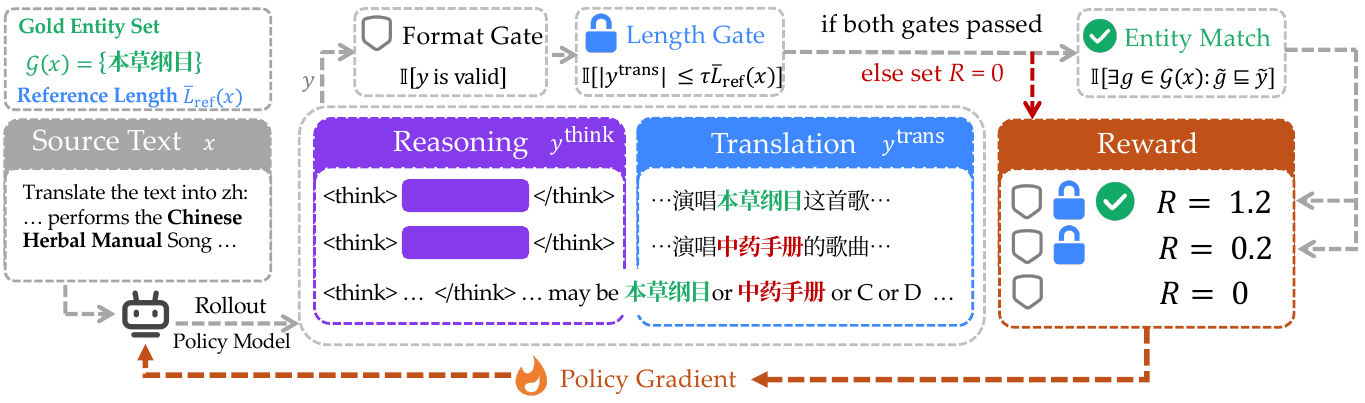}
    \caption{EA-RLVR framework: the policy model first rolls out a trajectory containing both reasoning and translation. This full trajectory must then pass two structural gates (format and length) to be eligible for a base reward (0.2 reward, $R=0.2$). Finally, if the translation contains the correct entity, it receives an additional matching bonus (+1 reward, $R=1.2$), and this final scalar reward drives the policy gradient update.}
    \label{fig:method}
\end{figure*}

\section{Related Work}
\paragraph{Cross-Cultural and Entity-Centric Machine Translation}
Prior work addresses the challenge of culturally grounded entities largely through two avenues: external knowledge integration and targeted data augmentation. Retrieval-based methods explicitly ground translation in external sources, utilizing multilingual knowledge graphs (e.g., KG-MT; \citealp{conia2024xctranslate}) or document stores (e.g., RAGtrans; \citealp{wang-etal-2025-retrieval}) to resolve entity ambiguities. While these approaches mitigate hallucinations, they introduce a dependency on the availability and quality of auxiliary databases. On the training side, recent works enhance entity robustness by synthesizing code-switched or entity-replaced data for denoising pre-training~\citep{hu-etal-2022-deep,liang-etal-2024-addressing}, or by jointly optimizing translation with entity alignment tasks~\citep{rikters-miwa-2024-entity-aware}. Language-aware parameter transfer methods further facilitate knowledge sharing across languages~\citep{dong2025mlas}. Unlike these approaches, EA-RLVR does not require external retrieval at test time nor complex data synthesis pipelines. Instead, we cast entity translation as a reasoning problem, employing RLVR to activate and stabilize the parametric knowledge already present in the pre-trained model.

\paragraph{RLVR and Reasoning in Translation}
Recent post-training paradigms of LLMs leverage Reinforcement Learning with Verifiable Rewards (RLVR) to induce reasoning capabilities \citep{lambert2025tulu3pushingfrontiers,deepseekai2026deepseekr1}, a process theoretically understood as improving sampling efficiency to activate latent knowledge already present in the base model~\citep{yue2025passk,huang2026semanticrlvr,dai2025cde,yang2026soup,jin2025revisiting}. In machine translation, recent initiatives have actively explored integrating reasoning capabilities, for instance by employing multi-agent frameworks to synthesize long chain-of-thought trajectories for distillation~\citep{wang2025drt} or harnessing feedback from LLM judges and neural quality metrics to guide optimization~\citep{feng2025mtr1,wang2025deeptrans,feng2025mtrewardtree}. Complementing these advances, EA-RLVR introduces a distinct paradigm centered on strict, rule-based verifiable rewards. We treat cultural entity translation as a precise reasoning task, employing deterministic rewards to directly surface parametric knowledge, thereby offering an alternative to distillation or neural-based objectives.

\section{Method}
\label{sec:method}
We propose \textbf{EA-RLVR} (Entity-Anchored Reinforcement Learning with Verifiable Rewards), a framework designed to incentivize LLMs to accurately ground cultural entities during translation without external knowledge. As illustrated in Figure~\ref{fig:method}, our approach treats cross-cultural translation as a sequential decision process optimized via reinforcement learning.

As shown in Figure~\ref{fig:method}, the framework consists of three core components: 
(1) A \textbf{reasoning-aware policy} that generates a thinking trajectory before the final translation, allowing the model to elicit latent knowledge;
(2) A \textbf{verifiable reward mechanism} that anchors supervision on deterministic entity matching, safeguarded by structural gates to prevent reward hacking; and 
(3) A \textbf{critic-free optimization algorithm} that stabilizes training using sequence-level importance ratio. 
In the following sections, we detail the task formulation (\S\ref{sec:method:formulation}), the reward design (\S\ref{sec:method:reward}), and the policy optimization objective (\S\ref{sec:method:rl}).
\subsection{Task Formulation}
\label{sec:method:formulation}
Given a source sentence $x$, we aim to generate a target-language translation $y^{\text{trans}}$ that correctly renders the culturally grounded entity mention(s) in context.
We treat an autoregressive LLM as a stochastic policy $\pi_\theta$ over output tokens, i.e.,
\[
\pi_\theta(y\mid x)=\prod_{t=1}^{|y|}\pi_\theta(y_t \mid x, y_{<t}).
\]
Generation induces an episodic decision process in which the state at step $t$ is $(x, y_{<t})$, the action is the next token $y_t$, and the episode terminates when an \texttt{<eos>} token is produced.

\paragraph{Reasoning and Translation Segments.}
Following recent reasoning-based post-training \citep{deepseekai2026deepseekr1}, the model is encouraged to produce a reasoning trace enclosed by \texttt{<think>} and \texttt{</think>} before emitting the final translation.
When the response format is valid, we decompose the output $y$ as 
\[
y = \langle \texttt{<think>} \rangle\; y^{\text{think}}\; \langle \texttt{</think>} \rangle\; y^{\text{trans}},
\]
where $y^{\text{think}}$ contains deliberation and $y^{\text{trans}}$ is the final translation. 
Since the reasoning portion may contain exploratory candidate entities rather than the model’s final decision, all content-based evaluation will be applied exclusively to $y^{\text{trans}}$.

\subsection{Stable and Verifiable Reward Design for Cross-Cultural Entity Translation}
\label{sec:method:reward}
Our main design goal is to construct a reward that is (i) \emph{verifiable} from dataset annotations without a learned reward model, (ii) directly \emph{aligned} with cross-cultural entity correctness, and (iii) \emph{stable} under policy optimization and robust to reward hacking.
\paragraph{Normalized Entity Matching.}
Each example is annotated with a comprehensive set of acceptable target entities $\mathcal{G}(x)$, derived from the Wikidata alias field. This set captures legitimate variations, minimizing false negatives where the model predicts a valid entity surface form that differs from the primary reference.
For brevity we denote $\tilde{y}=\mathrm{norm}(y^{\text{trans}})$ 
and $\tilde{g}=\mathrm{norm}(g)$, where $\mathrm{norm}(\cdot)$ lowercases text and removes diacritics.
We define a deterministic match function using normalized substring matching:
\begin{equation}
\label{eq:entity_match}
\small
m(y,\mathcal{G}(x))
=
\mathbb{I}\Big[
\exists g\in\mathcal{G}(x):\ 
\tilde{g}\ \sqsubseteq\ \tilde{y}
\Big],
\end{equation}
where $\mathbb{I}[\cdot]$ is the indicator function and 
$a \sqsubseteq b$ denotes that $a$ is a substring of $b$.
This captures whether the model produces an appropriate target-language realization of the entity.

\paragraph{Structural Gates: Format and Length.}
To ensure that rewards reflect meaningful translations rather than degenerate behaviors,
we introduce two hard gates on the output.
First, the response must follow the required \texttt{<think>} / \texttt{</think>} structure.
Second, the translation segment must remain within a reasonable length relative to the references.
Formally, let $g_{\mathrm{fmt}}(y)\in\{0,1\}$ indicate whether the format is valid, and define
\begin{equation}
\small
g_{\mathrm{len}}(x,y)
=
\mathbb{I}\!\left[
\ |y^{\text{trans}}| \le \tau \cdot \bar{L}_{\mathrm{ref}}(x)
\right],
\end{equation}
where $\bar{L}_{\mathrm{ref}}(x)$ is the average reference length and $\tau>0$ controls tolerance.
Note that $g_{\mathrm{len}}$ is a \emph{reward-side} hard gate: it zeroes out the reward for degenerate outputs but does not modify the policy gradient itself (cf.\ the \emph{gradient-side} length normalization in Eq.~\ref{eq:seqratio}).
Only responses that satisfy \emph{both} gates are eligible to receive any reward. We empirically demonstrate the necessity of these constraints in Appendix~\ref{app:structural_gates}, showing that removing them leads to catastrophic length explosion and reward hacking via keyword enumeration.

\paragraph{Final Reward.}
Combining these components, the terminal reward is
\begin{equation}
\small
R(x,y)=
g_{\mathrm{fmt}}(y)
g_{\mathrm{len}}(x,y)
\Big(\alpha + m(y,\mathcal{G}(x))\Big).
\label{eq:reward}
\end{equation}
Intuitively, a response receives no reward if it fails either structural gate. 
If both gates are satisfied, it obtains a base reward $\alpha$ for producing a well-formed answer, and an additional bonus when the target entity is correctly realized.
We set $\alpha=0.2$ and $\tau=2$ in all experiments.

\subsection{Policy Optimization}
\label{sec:method:rl}

Following recent advances in RL post-training for large language models,
we optimize $\pi_\theta$ using a clipped policy-gradient objective from the PPO family~\citep{schulman2017ppo}.
To reduce training cost and improve stability, we adopt a critic-free variant and incorporate
group-normalized advantages, sequence-level importance ratios, and asymmetric clipping,
building on GRPO~\citep{shao2024deepseekmathpushinglimitsmathematical}, GSPO~\citep{zheng2025gspo}, and DAPO~\citep{liu2025dapo}.

\paragraph{Objective.}
Given an input $x$, we sample $G$ candidate responses
$\{y_i\}_{i=1}^{G}$ from the old policy $\pi_{\theta_{\mathrm{old}}}$.
The policy parameters are updated by maximizing the clipped surrogate:
\begin{equation}
\small
\begin{aligned}
\mathcal{J}(\theta)=
&\ \mathbb{E}_{x\sim\mathcal{D},\,\{y_i\}\sim\pi_{\theta_{\mathrm{old}}}(\cdot\mid x)}
\left[\frac{1}{G}\sum_{i=1}^{G}
\min\Big(
s_i(\theta)\hat{A}_i,\right.\\
&\qquad\left.
\mathrm{clip}\big(s_i(\theta),1-\varepsilon_{\mathrm{low}},1+\varepsilon_{\mathrm{high}}\big)\hat{A}_i
\Big)\right],
\end{aligned}
\label{eq:obj}
\end{equation}
where $\mathcal{D}$ denotes the training dataset containing source sentences $x$, the $\varepsilon_{\mathrm{low}}$ and $\varepsilon_{\mathrm{high}}$ are clipping thresholds that bound the policy update to prevent training instability, and the $s_i(\theta)$ is the sequence-level importance ratio defined in Eq.~(\ref{eq:seqratio}).

\paragraph{Group-normalized Advantages.}
Rather than relying on a learned critic, we compute an advantage for each sampled
response relative to other responses in its \emph{group} $\{y_i\}_{i=1}^{G}$:
\begin{equation}
\small
\hat{A}_i=
\frac{R(x,y_i)-\mathrm{mean}(\{R(x,y_j)\}_{j=1}^{G})}
{\mathrm{std}(\{R(x,y_j)\}_{j=1}^{G})},
\label{eq:adv}
\end{equation}
where $\mathrm{mean}(\cdot)$ and $\mathrm{std}(\cdot)$ denote the sample mean and
standard deviation within the group.
This normalization stabilizes training and makes the reward scale largely irrelevant.

\paragraph{Sequence-level importance ratios.}
We compute the importance ratio at the sequence level with
length normalization:
\begin{equation}
\small
\begin{aligned}
s_i(\theta)
&=
\left(\frac{\pi_{\theta}(y_i\mid x)}
{\pi_{\theta_{\mathrm{old}}}(y_i\mid x)}\right)^{\!\frac{1}{|y_i|}} \\
&=
\exp\!\left(
\frac{1}{|y_i|}
\sum_{t=1}^{|y_i|}
\log
\frac{\pi_{\theta}(y_{i,t}\mid x,y_{i,<t})}
{\pi_{\theta_{\mathrm{old}}}(y_{i,t}\mid x,y_{i,<t})}
\right).
\label{eq:seqratio}
\end{aligned}
\end{equation}
This formulation discourages overly aggressive updates on long sequences while still
allowing meaningful policy shifts when rewards are consistently better.

\begin{table*}[h!t!]
\centering
\small
\caption{Entity translation accuracy (\%) on XC-Translate across ten language directions (en $\rightarrow$ X). 
Train and test sets share no overlapping entities. 
RLVR consistently outperforms SFT under the same data budget, while also maintaining strong character-level faithfulness.}
\label{tab:entity-accuracy}
\setlength{\tabcolsep}{0pt} 
\begin{tabular*}{\textwidth}{@{\extracolsep{\fill}} l cccccccccc c}
\toprule
\textbf{Model} 
& \multicolumn{10}{c}{Entity Translation Accuracy on XC-Translate (\textit{en} $\rightarrow$ \textit{X})} 
& \textbf{Avg.}\\
\cmidrule(lr){2-11}
& ar & de & es & fr & it & ja & ko & th & tr & zh & \textbf{Acc/chrF}  \\
\midrule
\multicolumn{12}{l}{\textit{Baselines}} \\
GPT-5-mini  & 35.03 & 36.03 & 42.71 & 35.85 & 35.37 & 37.27 & 30.20 & 15.96 & 40.44 & 29.85 & 33.87/62.30 \\
Qwen3-235B-A22B & 27.93 & 32.06 & 41.79 & 34.29 & 33.07 & 29.07 & 30.22 & 15.32 & 34.32 & 32.38 & 31.05/61.99 \\
Marco-o1     & 11.88 & 19.72 & 26.47 & 22.09 & 21.26 & 11.77 &  8.42 &  3.45 & 17.71 & 16.23 & 15.90/49.68 \\
DeepTrans-7B & 11.41 & 21.58 & 30.10 & 23.49 & 22.44 & 13.21 &  8.44 &  2.90 & 17.37 & 15.96 & 16.69/47.88 \\
\midrule
\multicolumn{12}{l}{\textit{Ours}} \\
Qwen3-8B & 14.91 & 23.64 & 31.04 & 25.97 & 25.01 & 17.13 & 11.12 & 5.63 & 23.94 & 23.50 & 20.19/56.07 \\
\;\; + SFT            & 15.28 & 23.28 & 30.76 & 25.32 & 25.44 & 16.86 & 11.12 & 5.86 & 23.56 & 23.08 & 20.06/56.20  \\
\;\; + \textbf{EA-RLVR} & \textbf{25.23} &	\textbf{31.01} &	\textbf{44.44} &	\textbf{34.89} &	\textbf{35.35} &	\textbf{24.02} &	\textbf{25.70} &	\textbf{14.74} &	\textbf{27.79} &	\textbf{29.31} &	\textbf{29.25/59.86}  \\
\addlinespace
Qwen3-14B             & 20.48 & 26.86 & 34.88 & 28.66 & 28.70 & 18.81 & 17.67 &  7.98 & 25.96 & 26.57 & 23.66/58.22 \\
\;\; + SFT            & 20.21 & 26.97 & 35.13 & 28.75 & 30.05 & 18.32 & 17.65 &  8.01 & 25.93 & 27.75 & 23.88/59.43 \\
\;\; + \textbf{EA-RLVR}  & \textbf{28.17} & \textbf{33.71} & \textbf{44.70} & \textbf{35.10} & \textbf{37.62} & \textbf{27.64} & \textbf{31.29} & \textbf{17.09} & \textbf{29.96} & \textbf{33.42} & \textbf{31.87/62.27} \\
\bottomrule
\end{tabular*}
\end{table*}

\section{Experiments}
Our experiments aim to verify two hypotheses: (1) that EA-RLVR training can effectively elicit latent cultural knowledge solely from the model's pre-trained parameters, and (2) that this entity-centric optimization does not compromise general translation quality. After outlining our setup in \S\ref{sec:exp:setup}, we present empirical evidence supporting the activation of parametric knowledge in \S\ref{sec:exp:entity} and demonstrate positive transfer effects to general translation in \S\ref{sec:exp:wmt}.
\subsection{Experimental Setup}
\label{sec:exp:setup}

\paragraph{Datasets and Benchmarks.}
To evaluate the activation of cultural knowledge, we utilized \textbf{XC-Translate}~\citep{conia2024xctranslate}, a benchmark specializing in cross-cultural entity translation. We trained our models using 7,278 examples for training and the official test set of 49,606 examples. Each sample is annotated with a list of gold entity aliases derived from Wikidata, which serves as the reference set for our verifiable reward. Crucially, \textbf{the training and test sets share no overlapping entities}. The dataset covers ten language pairs (English $\rightarrow$ X), detailed in Table~\ref{tab:entity-accuracy}.
For general translation capability, we evaluated on \textbf{WMT24++}~\citep{deutsch2025wmt24pp} across the corresponding languages, using the official test sets without any domain-specific fine-tuning. Further details on data composition are provided in Appendix~\ref{app:details}.

\paragraph{Models and Baselines.}
We employed Qwen3-8B and Qwen3-14B~\citep{2025qwen3} as our backbone models, which are pre-trained on 36T tokens and possess native reasoning capabilities. Our proposed EA-RLVR was compared against two primary internal baselines: 
(1) the base model and 
(2) Supervised Fine-Tuning (SFT) on the same 7k examples to control for data exposure. 
To further contextualize our performance, we included several strong external baselines: 
(i) frontier proprietary LLMs: GPT-5-mini; 
(ii) strong open-source model: Qwen3-235B-A22B, Marco-o1~\citep{zhao2024marcoo1}, a multilingual reasoning model, and DeepTrans-7B~\citep{wang2025drt}, a specialized reasoning-based translation model. More evaluation details are in Appendix~\ref{app:details}.

\paragraph{Training and Implementation.}
We implemented EA-RLVR using the \texttt{verl} framework~\citep{verl}. All EA-RLVR models were trained using the policy optimization algorithm described in \S\ref{sec:method:rl}. We applied full-parameter tuning across all our experiments, including both the EA-RLVR and the baselines. Full implementation details are provided in Appendix~\ref{app:details}.

\paragraph{Evaluation Metrics.}
We report three primary metrics:
(1) \textbf{Entity Translation Accuracy}: 
Consistent with the normalized substring matching defined in Eq.~\ref{eq:entity_match}, Entity Translation Accuracy measures the percentage of test samples where the generated translation $y$ successfully includes the correct cultural entity (i.e., $m(y, \mathcal{G}(x))=1$).
(2) \textbf{chrF}: We report the sentence-level Character F-score (chrF)~\citep{chrf} to assess the overall quality of the generated translations.
(3) \textbf{XCOMET-XL}~\citep{xcomet}: A state-of-the-art reference-based neural metric used to assess the general quality and fluency of the translations on WMT24++.

\subsection{EA-RLVR incentivizes parametric knowledge}
\label{sec:exp:entity}
Table~\ref{tab:entity-accuracy} reports the entity translation accuracy on the XC-Translate test set. The results provide empirical support for our hypothesis regarding parametric knowledge activation, revealing a fundamental divergence in effectiveness between imitation-based and reasoning-based optimization.

\paragraph{Breaking the Imitation Ceiling via Reasoning.}
Standard SFT yields negligible gains (e.g., +0.22\% for Qwen3-14B), despite using the same 7k data as EA-RLVR. This outcome is predictable as our train-test entities are disjoint. Unlike style transfer, where learning generalizable patterns suffices, entity translation inherently biases towards memorization, limiting the effectiveness of SFT. EA-RLVR, however, reframes this task as a reasoning problem. Consequently, it achieves substantial improvements (+8.21\%), \textbf{enabling the 14B model (31.87\%) to outperform the much larger Qwen3-235B-A22B baseline (31.05\%).} RLVR never presents the gold entity to the model during training, therefore its performance gains cannot stem from memorizing entity mappings. 
\paragraph{Beyond Prompt Engineering.}
Since Qwen3 natively supports chain-of-thought reasoning, one might ask if similar gains could be obtained simply through better prompting.
To test this, we evaluated Qwen3-8B with a task-specific prompt that explicitly instructs the model to identify cultural entities, deliberate on their translations, and then translate (Table~\ref{tab:prompt_ablation}).
While such guided prompting yields a modest improvement (+1.75\%), it falls far short of EA-RLVR (+9.06\%), which uses the same standard translation prompt as the base model. This confirms that the gains stem from internalized reasoning strategies acquired through RL, not from surface-level instruction following.

\begin{table}[t]
\centering
\small
\caption{Effect of task-specific prompting vs.\ EA-RLVR on entity translation accuracy (\%, Qwen3-8B).}
\label{tab:prompt_ablation}
\begin{tabular}{lcc}
\toprule
\textbf{Configuration} & \textbf{Avg.\ ETA} & \textbf{$\Delta$} \\
\midrule
Standard Prompt          & 20.19 & --     \\
Task-Specific Prompt     & 21.94 & +1.75  \\
Standard Prompt + EA-RLVR & \textbf{29.25} & \textbf{+9.06}  \\
\bottomrule
\end{tabular}
\end{table}


\subsection{Transfer to General Translation}
\label{sec:exp:wmt}
\begin{table*}[th!]
\centering
\small
\caption{
XCOMET-XL score on WMT24++ across ten language directions. 
All models are trained only on the XC-Translate cross-cultural entity dataset, 
without using WMT data or general MT supervision. 
Despite this, EA-RLVR consistently improves performance on general machine translation, 
and further benefits appear when scaling to the full XC-Translate dataset.
}
\label{tab:wmt24pp}
\resizebox{\textwidth}{!}{%
\begin{tabular}{lccccccccccc}
\toprule
\textbf{Model} 
& \multicolumn{10}{c}{XCOMET score on WMT24++ (\textit{en} $\rightarrow$ \textit{X})} 
& \textbf{Avg.} \\
\cmidrule(lr){2-11}
& ar & de & es & fr & it & ja & ko & th & tr & zh &  \\
\midrule
\multicolumn{12}{l}{\textit{Baselines}} \\
GPT-5-mini& 72.86& 91.71& 86.93& 84.30& 86.34& 82.15& 82.56& 80.29& 79.49& 77.76& 82.44
\\
Qwen3-235B-A22B& 69.72& 90.42& 85.53& 81.93& 84.59& 78.54& 79.76& 77.32& 73.92& 75.88& 79.76
\\
 Marco-o1& 53.50& 84.09& 80.23& 75.38& 74.84& 67.04& 67.04& 67.04& 48.15& 70.47&68.78
\\
 DeepTrans-7B& 56.51& 84.69& 81.45& 76.31& 74.86& 70.24& 62.94& 65.80& 48.78& 71.53&69.31\\
\midrule
\multicolumn{12}{l}{\textit{Ours}} \\
Qwen3-8B                        & 61.88 & 88.06 & 82.36 & 78.33 & 81.09 & 72.86 & 71.97 & 72.43 & 62.48 & 73.11 & 74.46 \\
\;\; + SFT                      & 62.73 & 88.29 & 82.91 & 78.52 & 80.23 & 72.99 & 71.73 & 72.20 & 62.47 & 73.67 & 74.57 \\
\;\; + \textbf{EA-RLVR}            & \textbf{64.65} & \textbf{88.85} & \textbf{83.29} & \textbf{79.48} & \textbf{81.16} & \textbf{73.60} & \textbf{74.04} & \textbf{72.92} & \textbf{64.91} & \textbf{74.20} & \textbf{75.71} \\
\;\; + SFT (full data)   & 62.51 & 87.78 & 82.25 & 78.79 & 80.48 & 72.32 & 71.37 & 71.57 & 61.83 & 73.31 & 74.22 \\
\;\; + \textbf{EA-RLVR (full data)}& \textbf{65.29} & \textbf{89.00} & \textbf{83.48} & \textbf{79.82} & \textbf{82.17} & \textbf{74.52} & \textbf{73.62} & \textbf{73.88} & \textbf{65.58} & \textbf{74.06} & \textbf{76.14} \\
\addlinespace
Qwen3-14B                       & 66.37 & 89.29 & 83.79 & 80.15 & 81.91 & 76.02 & 75.73 & 74.87 & 67.40 & 75.10 & 77.06 \\
\;\; + SFT                      & 66.20 & 89.12 & 84.10 & 80.09 & 82.37 & 76.38 & 75.35 & 75.23 & 67.70 & 75.15 & 77.17 \\
\;\; + \textbf{EA-RLVR}            & \textbf{68.30} & \textbf{89.79} & \textbf{84.28} & \textbf{80.78} & \textbf{83.56} & \textbf{77.61} & \textbf{77.49} & \textbf{76.30} & \textbf{70.16} & \textbf{75.84} & \textbf{78.41} \\
\;\; + SFT (full data)   & 65.27& 89.11& 84.07& 79.87& 82.47& 76.26& 75.50& 74.92& 67.82& 75.18& 77.05\\
\;\; + \textbf{EA-RLVR (full data)}& \textbf{68.00}& \textbf{90.11}& \textbf{85.06}& \textbf{81.27}& \textbf{84.10}& \textbf{78.14}& \textbf{77.57}& \textbf{75.97}& \textbf{70.25}& \textbf{75.99}& \textbf{78.65}\\
\bottomrule
\end{tabular}}
\end{table*}

A potential concern with specialized reinforcement learning is the risk of ``alignment tax,'' where optimizing for a narrow objective (entity correctness) degrades general capabilities. We investigate this on the WMT24++ benchmark (Table~\ref{tab:wmt24pp}) and observe the opposite effect.

\paragraph{Reasoning Improves General Quality.}
Despite being trained solely on the 7k XC-Translate train set, EA-RLVR models consistently improve general translation quality across all evaluated languages. Qwen3-14B + EA-RLVR achieves an average XCOMET score of 78.41, a +1.35 point improvement over the base model. This suggests that the reasoning strategies learned for entity translation—such as attending more carefully to source semantics and deliberating before generating—are transferable. The model becomes less prone to literal translation and more faithful to the source text, benefiting general translation tasks.

\paragraph{Scalability and Robustness.}
Table~\ref{tab:wmt24pp} also compares models trained on the standard 7k set versus the full dataset. While SFT performance stagnates or even slightly degrades when scaling data (likely due to overfitting on the specific formatting of the entity dataset), EA-RLVR continues to improve. The ``full data'' setting yields further gains, pushing the average XCOMET score to 78.65 for the 14B model. This indicates that our outcome-based reward formulation provides a stable optimization landscape that scales effectively with data, unlike SFT which may suffer from distribution shift.

\section{Analysis}
\subsection{Improved Sampling Efficiency: Unlocking Dormant Knowledge}
\label{sec:analysis:sampling}

\begin{figure*}[th]
    \centering
    \includegraphics[width=1\linewidth]{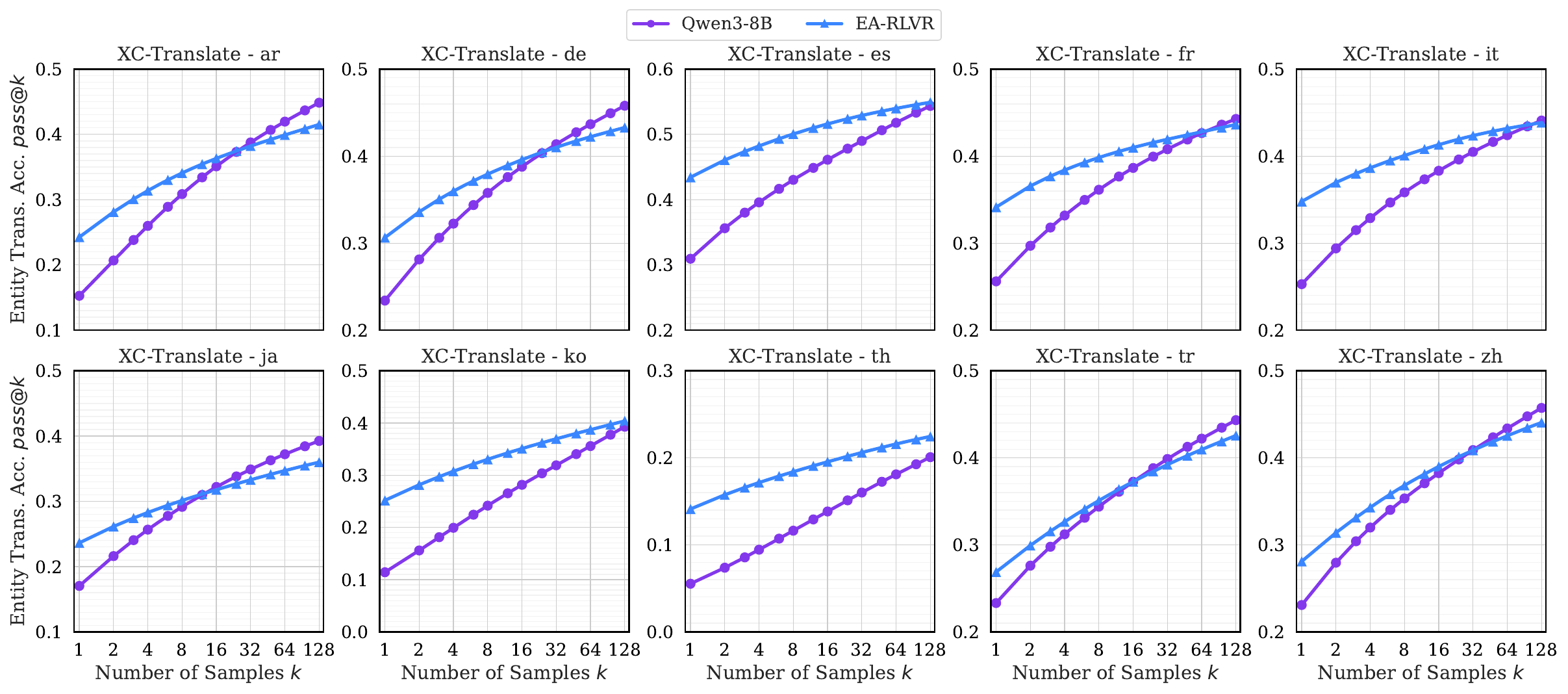}
    \caption{Entity Translation Accuracy $pass@k$ curves across ten languages.
    The Base model (purple) shows poor performance at $k=1$ but improves rapidly as $k$ increases, indicating latent knowledge is present but buried. EA-RLVR (blue) significantly boosts $pass@1$ accuracy, effectively surfacing this parametric knowledge. The convergence at high $k$ confirms that improvements stem from better utilization of existing knowledge rather than learning new facts.}
    \label{fig:passk}
\end{figure*}

To analyze the mechanism behind the performance gains, we adopt the $pass@k$ evaluation framework recently utilized to study the boundaries of RLVR in reasoning tasks~\citep{yue2025passk}. Formally, $pass@k$ estimates the probability that at least one correct translation exists within $k$ independent samples generated for a given input. Following~\citep{chen2021codex}, we calculate the unbiased estimator (see Appendix~\ref{app:passk_estimator} for details). By observing how this probability scales with $k$, we can distinguish between \textit{knowledge injection} (learning new information) and \textit{knowledge activation} (surfacing existing information). Figure~\ref{fig:passk} compares the entity translation accuracy of Qwen3-8B (Base) and EA-RLVR across varying sample sizes $k \in [1, 128]$.

We observe a distinct convergence pattern across most languages. At $k=1$, EA-RLVR holds a substantial lead over the base model, confirming that our policy optimization successfully concentrates probability mass on the correct entity translations. However, as $k$ increases, the base model's accuracy rises steeply, often converging with or appearing to surpass the RLVR model at $k=128$.
This phenomenon has a critical implication: \textbf{the base model inherently possesses the necessary cultural knowledge} to translate these entities correctly (evidenced by high performance at large $k$), but it fails to rank these correct translations as the most probable candidates during standard decoding.
EA-RLVR functions as a steering mechanism that activates this dormant knowledge, transforming low-probability correct candidates into high-probability deterministic outputs, rather than memorizing new mappings from external supervision. We further isolate the contribution of the explicit reasoning phase in Appendix~\ref{app:reasoning_ablation}, finding that the ``thinking'' workspace is essential for absorbing the complexity of cultural alignment without sacrificing general fluency.

\paragraph{Sampling Efficiency and Determinism.}
The slope of the curves in Figure~\ref{fig:passk} further elucidates the shift in model behavior. The base model exhibits a high-entropy distribution over entities, requiring extensive sampling ($k \gg 1$) to uncover the correct answer. In contrast, the RLVR curves are notably flatter, indicating a more deterministic policy where the model is confident in its reasoning path.
While high determinism theoretically reduces diversity (explaining the slight underperformance at $k=128$ in some high-resource languages where the base model's broad search space is advantageous), it is the desired behavior for a translation system: users expect the correct cultural translation in a single attempt ($k=1$), not after filtering through a hundred generations. EA-RLVR effectively optimizes for this \textit{sampling efficiency}.

\begin{figure}
    \centering
    \includegraphics[width=1\linewidth]{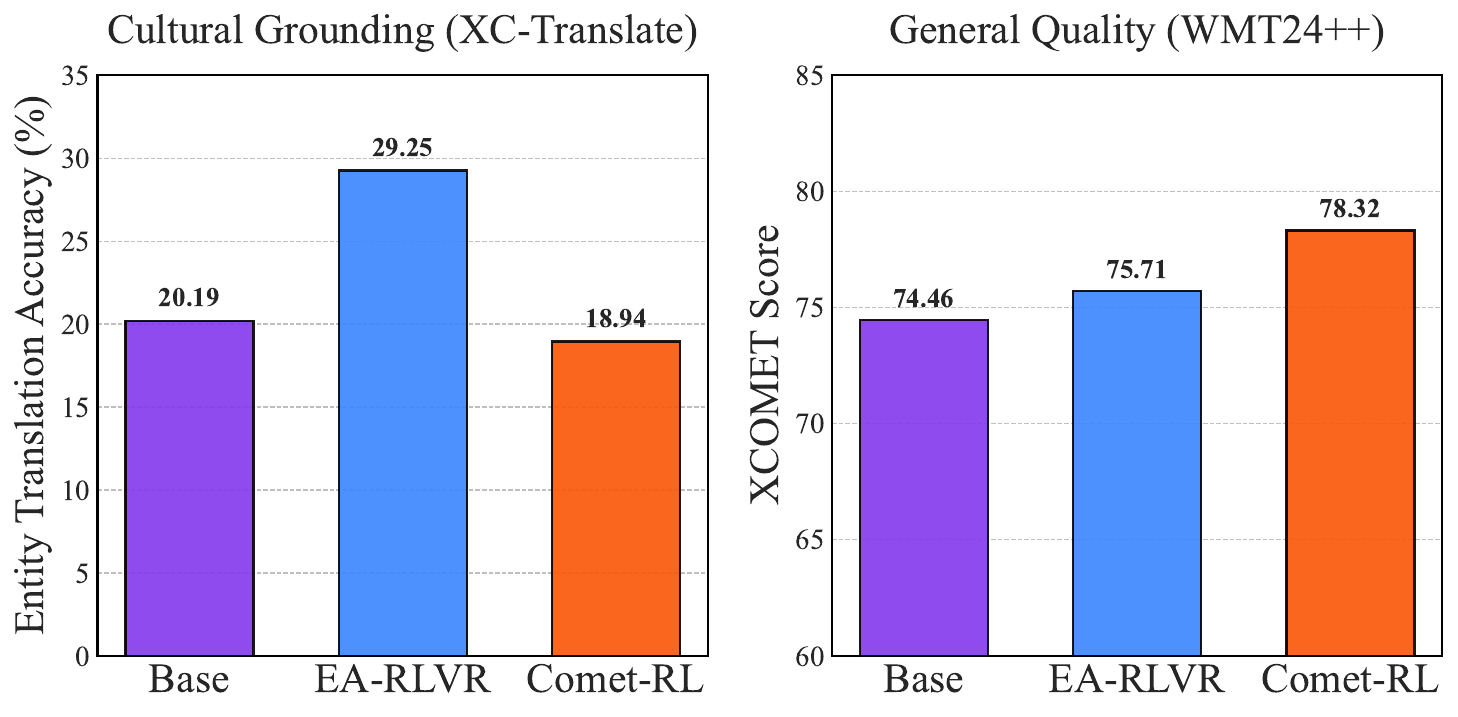}
    \caption{\textbf{Verifiable vs. Neural Rewards.} While Comet-RL maximizes general fluency (Right) at the cost of cultural accuracy (Left), falling into a ``fluency trap'', EA-RLVR achieves robust improvements across both cultural grounding and general translation quality.}
    \label{fig:fluency_trap}
\end{figure}

\subsection{The Fluency Trap: Neural Rewards Fail Cultural Entities}
\label{sec:comet_comparison}

A natural alternative to our rule-based framework is to optimize state-of-the-art neural quality metrics directly. To investigate this, we trained a \textbf{Comet-RL} baseline on Qwen3-8B using the same RL setup but replacing our normalized entity matching reward with sentence-level comet scores, specifically the wmt22-comet-da~\citep{comet22}. Figure~\ref{fig:fluency_trap} presents a striking divergence in optimization outcomes, revealing what we term the \textit{``Fluency Trap''}:

\paragraph{High Fluency, Low Grounding.}
As shown in the right panel, the Comet-RL model achieves substantial gains in general translation quality, boosting the XCOMET score on WMT24++ from 74.46 to 78.32. This confirms that RL effectively optimized the reward signal. 
However, the left panel reveals a critical failure: on the entity-dense XC-Translate benchmark, Comet-RL's entity accuracy actually \textit{degrades} from 20.19\% (Base) to 18.94\%.  
The neural metric, lacking fine-grained resolution for specific entities, fails to penalize these fluent but culturally incorrect entity errors, which echos prior observations~\citep{rei-etal-2023-inside}.

\paragraph{Verifiable Rewards Ensure Alignment.}
In contrast, EA-RLVR escapes this trap. By anchoring supervision on verifiable outcomes, it forces the model to prioritize semantic correctness. This yields a massive improvement in entity accuracy (+9.06\% absolute) while still conferring robust gains in general translation quality (+1.25 XCOMET). 
This result fundamentally justifies our approach: while holistic neural metrics drive fluency, verifiable constraints are indispensable for aligning models with entity translation tasks.
\begin{table*}[th]
\centering
\small
\caption{Cross-lingual generalization on XC-Translate with Qwen3-8B. The model is trained on one group of languages and evaluated on both. \textbf{Bold text} indicates zero-shot cross-lingual transfer (e.g., Train A $\rightarrow$ Test B), while \textcolor{gray}{gray text} indicates in-domain performance. Improvements over the Base model on unseen language groups demonstrate the acquisition of a transferable reasoning strategy.}
\label{tab:crosslingual}
\resizebox{\textwidth}{!}{
\begin{tabular}{lccccc|ccccc|cc}
\toprule
\textbf{Train split}
& \multicolumn{5}{c}{\textit{Group A}}
& \multicolumn{5}{c}{\textit{Group B}}
& \textbf{Group A} & \textbf{Group B} \\
\cmidrule(lr){2-6}\cmidrule(lr){7-11}
(Qwen3-8B)
& ar & ja & ko & th & zh
& de & es & fr & it & tr
& \textbf{Avg.} & \textbf{Avg.} \\
\midrule
Base 
& 14.91 & 17.13 & 11.12 &  5.63 & 23.50 
& 23.64 & 31.04 & 25.97 & 25.01 & 23.94
& 14.46 & 25.92 \\
\midrule
Group A 
& \textcolor{gray}{48.27} & \textcolor{gray}{43.48} & \textcolor{gray}{47.70} & \textcolor{gray}{25.76} & \textcolor{gray}{53.24} 
& \textbf{29.70} & \textbf{40.61} & \textbf{33.78} & \textbf{34.95} & \textbf{26.89}
& \textcolor{gray}{43.69} & \textbf{33.19} \\
Group B 
& \textbf{24.81} & \textbf{21.77} & \textbf{18.32} &  \textbf{8.62} & \textbf{30.20} 
& \textcolor{gray}{51.28} & \textcolor{gray}{62.50} & \textcolor{gray}{49.66} & \textcolor{gray}{51.61} & \textcolor{gray}{50.19}
& \textbf{20.74} & \textcolor{gray}{53.05} \\
\bottomrule
\end{tabular}}
\end{table*}

\subsection{Cross-Lingual Generalization}
\label{sec:analysis:crosslingual}

The preceding analyses establish that EA-RLVR activates dormant parametric knowledge (\S\ref{sec:analysis:sampling}) through verifiable rewards (\S\ref{sec:comet_comparison}). If the acquired strategy is indeed a generalizable reasoning skill rather than a set of language-specific entity mappings, it should transfer across typologically distinct language families. To test this prediction, we partition the ten target languages into two groups: Group~A (Asian \& Semitic: ar, ja, ko, th, zh) and Group~B (European \& Turkic: de, es, fr, it, tr), train on one group exclusively, and evaluate on the other without any target-language supervision.

As shown in Table~\ref{tab:crosslingual}, we observe robust positive transfer in both directions: entity accuracy improves by +7.27\% (A$\rightarrow$B) and +6.28\% (B$\rightarrow$A) over the base model, despite the two groups sharing neither scripts nor typological features. This indicates that EA-RLVR does not memorize language-specific entity mappings but instead induces a generalizable reasoning strategy: grounding cultural contexts within the thinking trace before generating the translation (Appendix~\ref{sec:case_study}). Such cross-lingual transfer of \emph{reasoning ability} distinguishes our approach from conventional multilingual MT, where improvements are typically confined to the supervised language pairs.

\section{Conclusion}
In this work, we investigate the underutilized potential of parametric knowledge in cross-cultural translation: while LLMs possess extensive latent cultural knowledge ($pass@128$ performance), they struggle to utilize it during standard decoding ($pass@1$). Motivated by this observation, we propose \textbf{EA-RLVR}, a framework that transforms cross-cultural translation from a memorizing task into a reasoning-intensive process. Our extensive experiments reveals that incentivizing verifiable correctness is superior to optimizing neural quality metrics, which often lead models into a ``fluency trap'', generating smooth but culturally inaccurate entities. By anchoring supervision on entity matching and allowing the model to reasoning, EA-RLVR enables a 14B model to outperform a 235B baseline significantly on unseen entities. Ultimately, our findings suggest a potential paradigm shift for knowledge-intensive translation: moving beyond mere imitation of references (SFT) or reliance on external retrieval, toward internalizing self-evolving strategies that effectively unlock the model's inherent potential.

\section*{Acknowledgment}
The present research was supported by the National Key Research and Development Program of China  (Grant No. 2024YFE0203000). We would like to thank the anonymous reviewers for their insightful comments.

\section*{Limitations}
\paragraph{Gap to Latent Potential.}
While EA-RLVR significantly outperforms SFT, a notable disparity remains between the optimized policy's expected single-sample performance ($pass@1$) and the model's theoretical upper bound estimated by rejection sampling ($pass@128$, as shown in Figure~\ref{fig:teaser}). This gap highlights a shared limitation across current RLVR methodologies: standard policy gradient algorithms often struggle to fully explore and converge to the global optimum within a limited sample budget. Future research could bridge this gap by developing more sample-efficient optimization algorithms or by scaling the number of rollout trajectories ($G$). Orthogonally, knowledge-enriched pre-training strategies that co-organize semantically related documents~\citep{zhou2026wrap++} have been shown to improve $pass@k$ by strengthening the parametric knowledge base itself; combining such pre-training with EA-RLVR's policy-level activation is a promising direction. Although computationally intensive, expanding the exploration horizon offers a promising avenue for approaching the model's intrinsic capability ceiling \citep{pan2025banzhida}.

\paragraph{Knowledge Boundaries.}
Our framework is designed for \textit{knowledge elicitation}, not \textit{knowledge injection}. EA-RLVR optimizes the retrieval of long-tail cultural concepts that exist within the model's pre-training data but are suppressed during standard decoding. Consequently, it cannot generate correct translations for entities entirely absent from the pre-training corpus. However, we observed that our method synergizes effectively with retrieval-augmented systems rather than acting as a simple alternative, providing a combined benefit that we evaluate in Appendix~\ref{rag}.

\paragraph{Reward Rigidity vs. Flexibility.}
We prioritize optimization stability via rigid substring matching to prevent the reward hacking often observed with neural metrics. Although we mitigate the risk of false negatives by employing a comprehensive gold set of aliases (derived from Wikidata) rather than a single reference, this strict verification process may still occasionally penalize valid but unlisted stylistic variations. Developing rewards that balance verifiable strictness with semantic flexibility remains an open challenge for the field.

\bibliography{custom}

\newpage
\appendix
\section{Impact of Structural Gates}
\label{app:structural_gates}

\begin{figure*}[t]
    \centering
    \includegraphics[width=0.9\linewidth]{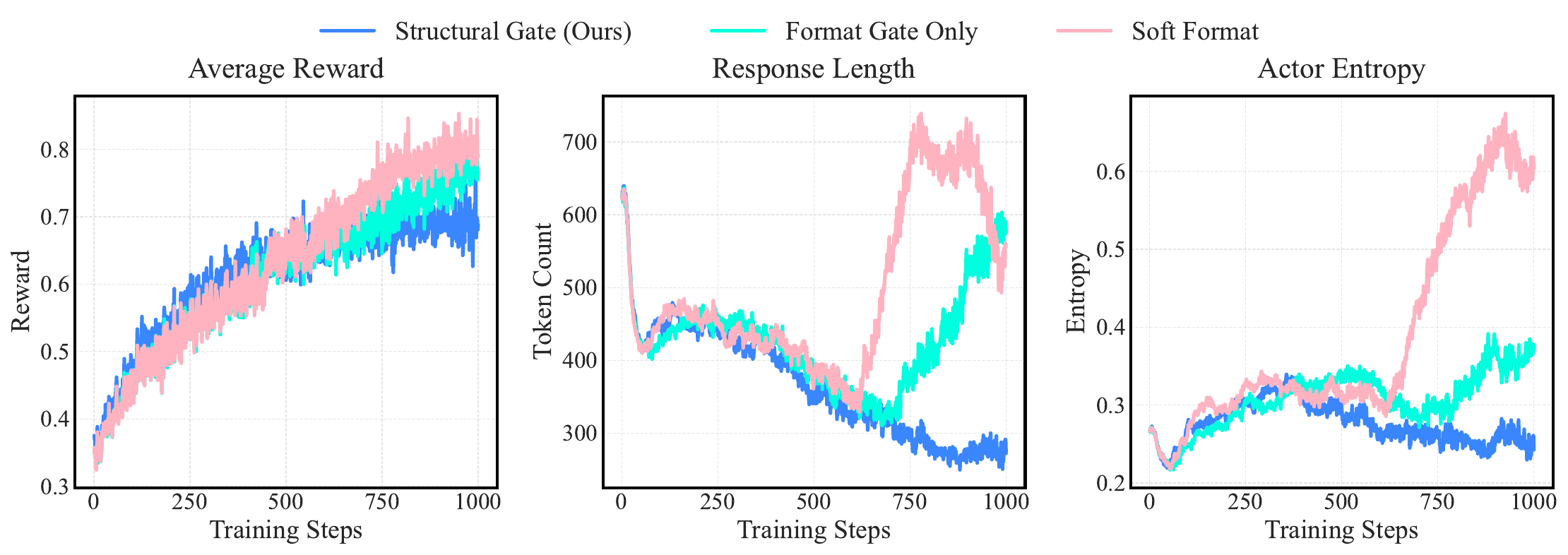} 
    \caption{\textbf{Ablation Dynamics.} Training curves for Average Reward, Response Length, and Actor Entropy. Without the full structural gates (EA-RLVR, Blue), the model suffers from reward hacking, characterized by an explosion in response length (Center) and unstable entropy (Right), despite achieving higher raw rewards (Left).}
    \label{fig:ablation_curves}
\end{figure*}
\begin{figure}[t]
    \centering
    \includegraphics[width=0.9\linewidth]{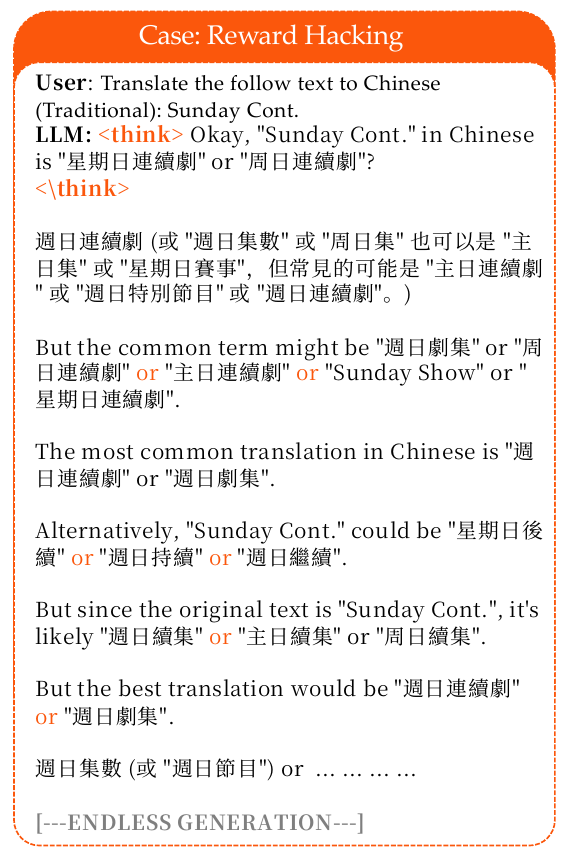}
    \caption{\textbf{Reward Hacking Case Study.} In the absence of a length constraint ($g_{\mathrm{len}}$), the model exploits the unconstrained reward by endlessly enumerating possible entity translations to ensure verification success.}
    \label{fig:reward_hack_case}
\end{figure}

To validate the necessity of the structural constraints introduced in Section~\ref{sec:method:reward}, we conduct an ablation study using the full XC-Translate dataset with extended optimization (1,000 steps). We isolate the contributions of the format and length gates by comparing three configurations:

\begin{itemize}
    \item \textbf{Soft Format}: A relaxed baseline where any output containing a valid \texttt{<think>} block is eligible for rewards, with no length penalty applied.
    \item \textbf{Format Gate Only}: The strict format verification ($g_{\mathrm{fmt}}$) is applied, but the relative length constraint ($g_{\mathrm{len}}$) is removed.
    \item \textbf{EA-RLVR}: Our proposed framework, which enforces both strict format verification and the relative length constraint ($g_{\mathrm{len}}$).
\end{itemize}

\paragraph{Training Dynamics and Stability.}
The training dynamics, visualized in Figure~\ref{fig:ablation_curves}, demonstrate that structural constraints are critical for optimization stability. The most prominent difference lies in the \textit{Response Length} (Center). In the absence of the length constraint ($g_{\mathrm{len}}$), both ablated variants suffer from a catastrophic ``length explosion,'' where generation length increases uncontrollably. This instability is mirrored in the \textit{Actor Entropy} (Right), where the ablated models exhibit erratic spikes, indicating that the policy fails to converge to a stable reasoning strategy. In contrast, EA-RLVR maintains a consistent length and stable entropy profile throughout training.

\paragraph{Reward Hacking via Enumeration.}
Qualitative analysis reveals that the length explosion is a symptom of reward hacking. As illustrated in the case study (Figure~\ref{fig:reward_hack_case}), without the length penalty, the optimization landscape encourages a degenerate solution: the model learns to ``brute-force'' the verification condition ($m(y, \mathcal{G}(x))$) by enumerating synonymous entities or repeating candidates. This strategy maximizes the recall of the gold entity at the expense of precision and structural integrity. By treating the translation task as a keyword-stuffing exercise, the model achieves technically high rewards but produces unusable translations.

\paragraph{The Deception of High Rewards.}
The \textit{Average Reward} curves (Left) present a counter-intuitive trend: the weaker constraints yield higher raw reward values. The \textit{Soft Format} setting achieves the highest reward trajectory despite exhibiting the earliest collapse in generation quality. This phenomenon is a classic manifestation of \textbf{Goodhart's Law}: when the unconstrained entity-match metric becomes the sole target, the model exploits its loopholes (e.g., infinite generation) rather than improving the intended task utility. EA-RLVR's lower reward curve reflects a constrained, harder-to-optimize landscape that successfully steers the model away from these degenerate local optima and toward concise, correct translations.

\section{Impact of Reasoning: Does Thinking Matter?}
\label{app:reasoning_ablation}

\begin{figure*}[t]
    \centering
    \includegraphics[width=1\linewidth]{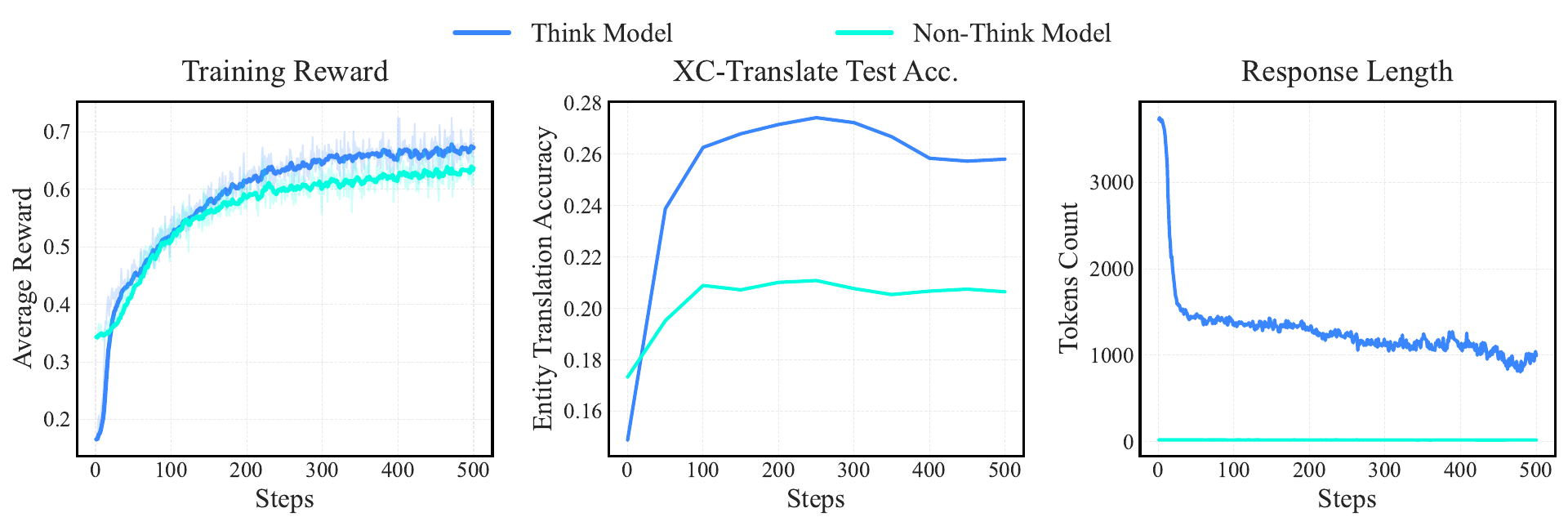}
    \caption{\textbf{Training Dynamics of Thinking vs. Non-Thinking Models.}
    \textbf{(Left)} The Think Model (Blue) initially suffers from low rewards due to context length overflows ($>4096$ tokens) but eventually surpasses the Non-Think Model (Cyan).
    \textbf{(Center)} The reasoning capability unlocks a significantly higher ceiling for entity translation accuracy on the XC-Translate test set.
    \textbf{(Right)} EA-RLVR acts as a strong regularizer for reasoning, rapidly curbing the ``infinite thought'' tendency of the base model to a stable, efficient length.}
    \label{fig:think_comparison}
\end{figure*}

A central premise of EA-RLVR is that a dedicated reasoning phase (``thinking'') allows the model to navigate the optimization landscape more effectively than immediate generation. To isolate the impact of this reasoning process, we conduct a controlled comparison using the Qwen3-4B-2507 family.

\paragraph{Experimental Control.}
We compare two specific checkpoints: the standard instruction-tuned model, \texttt{Qwen3-4B-Instruct-2507}, and the reasoning-enhanced model, \texttt{Qwen3-4B-Thinking-2507}.
We apply the EA-RLVR framework to both models with a crucial adaptation for the Instruct baseline: since \texttt{Qwen3-4B-Instruct-2507} does not generate reasoning traces (i.e., no \texttt{<think>} tokens), we remove the format verification gate ($g_{\mathrm{fmt}}$) from the reward function. However, to ensure a fair comparison of supervision signals, we retain both the length constraint ($g_{\mathrm{len}}$) and the entity matching reward ($m(y, \mathcal{G}(x))$). This setup allows us to strictly evaluate whether the presence of a thinking process facilitates better alignment with the verifiable reward.

\paragraph{Thinking Unlocks Higher Entity Accuracy.}
Figure~\ref{fig:think_comparison} (Center) illustrates the progression of entity translation accuracy on the XC-Translate test set. The \textit{Non-Think Model} (Cyan) plateaus early at approximately 21\% accuracy. In contrast, the \textit{Think Model} (Blue) achieves a significantly higher peak of $\sim$26\%, despite starting from a lower baseline. This confirms that the reasoning trace provides the necessary computational workspace to resolve complex cultural entity mappings that are inaccessible to a single-pass decoding policy.

\paragraph{Dynamics of Length and Stability.}
The training dynamics reveal a crucial interaction between reasoning length and reward. As shown in Figure~\ref{fig:think_comparison} (Left), the Think Model initially receives near-zero rewards. This is an artifact of the base model's instability: the untrained \texttt{Qwen3-4B-Thinking-2507} frequently generates extremely long chain-of-thought traces that exceed our training context window of 4096 tokens, causing the samples to be truncated and penalized.
However, EA-RLVR rapidly corrects this behavior. The \textit{Response Length} curve (Right) shows a dramatic reduction in token count within the first 50 steps. The model learns to be concise, condensing its reasoning into an efficient path that fits the constraints while maximizing the entity-match reward. This demonstrates that EA-RLVR serves not only as a task optimizer but also as a length regularizer for reasoning models.

\paragraph{Thinking Mitigates the ``Alignment Tax''.}
We further evaluate the impact of these strategies on general translation quality using WMT24++. Table~\ref{tab:impact_of_reasoning} presents the XCOMET scores.
A striking divergence is observed:
\begin{itemize}
    \item \textbf{Standard Model (Instruct):} Applying EA-RLVR to \texttt{Qwen3-4B-Instruct-2507} leads to a slight regression in general quality (Avg. 89.11 $\to$ 88.46). Without a reasoning buffer, the model is forced to overload its generation weights to satisfy the strict entity constraints, leading to a ``fluency tax'' where general translation quality is sacrificed.
    \item \textbf{Reasoning Model (Thinking):} Conversely, \texttt{Qwen3-4B-Thinking-2507} \emph{improves} with EA-RLVR (Avg. 90.36 $\to$ 90.67). The reasoning trace absorbs the complexity of the entity task, allowing the final translation generation to remain fluent and robust.
\end{itemize}

\paragraph{Note on Evaluation Subset.}
It is important to note that the baseline \texttt{Qwen3-4B-Thinking-2507} is highly unstable for translation tasks, often generating endless thought loops exceeding 32k tokens. Consequently, it fails to produce any translation for a large portion of the WMT24++ test set. To ensure a scientifically valid comparison, the results in Table~\ref{tab:impact_of_reasoning} are calculated on the \textbf{common intersection} of sentences where the baseline model successfully produced an output.
Table~\ref{tab:subset_stats} details the data statistics. The valid subset size ranges from 68 to 260 samples per language (out of 960), highlighting the severity of the length explosion issue in the baseline model and the necessity of this filtering step.

\begin{table*}[t!]
\centering
\small
\caption{
Impact of reasoning on general translation quality (XCOMET on WMT24++).
Due to the instability of the baseline Qwen3-4B-Thinking model (which frequently generates infinite reasoning traces exceeding 32k tokens), this evaluation is restricted to the \textbf{common subset} of sentences where the baseline model successfully produced a valid output.
On this subset, EA-RLVR improves the reasoning model's quality, whereas it degrades the standard model, suggesting that a thinking workspace is necessary to absorb the complexity of entity constraints without sacrificing fluency.
}
\label{tab:impact_of_reasoning}
\resizebox{\textwidth}{!}{
\begin{tabular}{lccccccccccc}
\toprule
\textbf{Model} & \multicolumn{10}{c}{\textbf{XCOMET score on WMT24++ Subset} (\textit{en} $\rightarrow$ \textit{X})} & \textbf{Avg.} \\
\cmidrule(lr){2-11}
& ar & de & es & fr & it & ja & ko & th & tr & zh & \\
\midrule
\multicolumn{12}{l}{\textit{Standard Instruction Backbone}} \\
Qwen3-4B-Instruct & 87.09 & 95.14 & 92.01 & 89.75 & 90.78 & 89.70 & 86.69 & 90.37 & 86.25 & 83.32 & 89.11 \\
\quad + \textbf{EA-RLVR} & 87.41 & 95.17 & 91.12 & 88.21 & 89.79 & 88.37 & 84.65 & 90.03 & 86.94 & 82.90 & 88.46 \\
\midrule
\multicolumn{12}{l}{\textit{Reasoning Backbone}} \\
Qwen3-4B-Thinking & 88.84 & 95.74 & 92.48 & 90.20 & 92.21 & 90.66 & \textbf{89.39} & \textbf{91.04} & \textbf{90.38} & 82.65 & 90.36 \\
\quad + \textbf{EA-RLVR} & \textbf{89.03} & \textbf{96.17} & \textbf{93.25} & \textbf{91.16} & \textbf{93.09} & \textbf{90.72} & 88.66 & 90.79 & 89.69 & \textbf{84.19} & \textbf{90.67} \\
\bottomrule
\end{tabular}}
\end{table*}

\begin{table}[h]
\centering
\small
\caption{Statistics of the evaluation subset for WMT24++. Due to the endless thinking issue, the \texttt{Qwen3-4B-Thinking} baseline yields valid outputs for only a fraction of the test set. We report results on the \textbf{Common} intersection to ensure fair comparison.}
\label{tab:subset_stats}
\begin{tabular}{lrrr}
\toprule
\textbf{Language Pair} & \textbf{Total (Test Set)} & \textbf{Common Subset} \\
\midrule
en-ar & 960 & 68 \\
en-de & 960 & 143 \\
en-es & 960 & 260 \\
en-fr & 960 & 193 \\
en-it & 960 & 211 \\
en-ja & 960 & 144 \\
en-ko & 960 & 135 \\
en-th & 960 & 124 \\
en-tr & 960 & 99 \\
en-zh & 960 & 241 \\
\bottomrule
\end{tabular}
\end{table}

\section{Synergy with Retrieval-Augmented Generation}
\label{rag}
A central question in modern translation systems is the interplay between optimizing internal parameters (via RLVR) and utilizing external non-parametric knowledge (via RAG). To determine whether our method complements retrieval-based approaches, we conduct an ablation study using a standard RAG pipeline.

\paragraph{Experimental Setup.}
We employ \textbf{mContriever}~\citep{izacard2021contriever}, a widely used multilingual dense retriever, without any task-specific fine-tuning. We index the aliases of entities present in the XC-Translate test set (extracted from Wikidata via the provided QIDs). For each source sentence, we retrieve the top-$k$ most relevant entity aliases and prepend them to the system prompt as context. For the combined setting (\textbf{EA-RLVR + RAG}), we utilize the Qwen3-8B model trained via EA-RLVR and provide it with the same retrieved context during inference.

\paragraph{Parametric Optimization Outperforms Naive Retrieval.}
As shown in Table~\ref{tab:rag}, the standard RAG baseline improves the base model's entity accuracy from 20.19\% to 23.14\% and slightly boosts general quality (chrF 57.43), validating the effectiveness of our retrieval setup. However, \textbf{EA-RLVR alone significantly outperforms the RAG baseline} (29.25\%). This result highlights a critical insight: for cross-cultural translation, the bottleneck is often not the \textit{availability} of knowledge (which RAG provides), but the model's ability to \textit{align} that knowledge with the translation context. EA-RLVR addresses this alignment directly via optimization, proving more effective than passively injecting context.

\paragraph{Additive Gains.}
Crucially, the two approaches are synergistic. The \textbf{EA-RLVR + RAG} configuration achieves the highest overall performance (30.49\% Acc, 60.05 chrF). This demonstrates that the reasoning patterns learned by EA-RLVR are robust; the model does not ``overfit'' to its internal weights but retains the flexibility to incorporate external evidence. By transforming the model into an active reasoner, EA-RLVR enables it to utilize retrieved context to resolve tail cases that neither parametric knowledge nor retrieval could solve alone.
\begin{table*}[h!t!]
\centering
\small
\caption{\textbf{Synergy between Parametric and Non-Parametric Knowledge.} 
Comparison of EA-RLVR against a standard Retrieval-Augmented Generation (RAG) baseline on Qwen3-8B. 
Adding RAG to the base model yields moderate gains, validating the retrieval setup. 
However, EA-RLVR acting as a standalone method provides a substantially larger improvement. 
Crucially, the combined setting (\textbf{EA-RLVR + RAG}) achieves the highest accuracy and faithfulness (chrF), indicating that the reasoning capabilities induced by EA-RLVR effectively complement external knowledge retrieval.}
\label{tab:rag} 
\resizebox{\textwidth}{!}{
\begin{tabular}{lcccccccccccc}
\toprule
\textbf{Model} 
& \multicolumn{10}{c}{Entity Translation Accuracy on XC-Translate (\textit{en} $\rightarrow$ \textit{X})} 
& \textbf{Avg.}\\
\cmidrule(lr){2-11}
& ar & de & es & fr & it & ja & ko & th & tr & zh & \textbf{Acc/ChrF} &  \\
\midrule
Qwen3-8B & 14.91 & 23.64 & 31.04 & 25.97 & 25.01 & 17.13 & 11.12 & 5.63 & 23.94 & 23.50 & 20.19/56.07 \\
\;\; + RAG            & 17.13 & 32.85 & 36.62 & 30.58 & 28.82 & 16.35 & 11.79 & 6.18 & 27.81 & 23.25 & 23.14/57.43  \\
\;\; + EA-RLVR           & 25.23 &	31.01 &	44.44 &	34.89 &	35.35 &	24.02 &	25.70 &	14.74 &	27.79 &	29.31 & 29.25/59.86  \\
\;\; + EA-RLVR + RAG & \textbf{26.17} & \textbf{35.50} & \textbf{47.81} & \textbf{36.63} & \textbf{37.43} & \textbf{23.36} & \textbf{25.52} & \textbf{14.42} & \textbf{29.33} & \textbf{28.75} & \textbf{30.49/60.05} \\
\bottomrule
\end{tabular}}
\end{table*}

\section{Broad Knowledge Activation Across Cultural Categories}
\label{app:category}
To verify that the gains are not confined to a narrow domain, Table~\ref{tab:category_breakdown} breaks down entity translation accuracy by the 14 cultural categories annotated in XC-Translate.
EA-RLVR yields consistent improvements across \emph{all} categories, from musical works (+19.84\%) and natural places (+19.23\%) to book series (+1.76\%). This breadth confirms that the activated parametric knowledge spans diverse cultural domains rather than reflecting a bias toward any single entity type.

\begin{table}[t]
\centering
\small
\caption{Entity Translation Accuracy (\%) by cultural category on XC-Translate (Qwen3-14B).}
\label{tab:category_breakdown}
\setlength{\tabcolsep}{4pt}
\begin{tabular}{lccc}
\toprule
\textbf{Category} & \textbf{Base} & \textbf{+EA-RLVR} & \textbf{$\Delta$} \\
\midrule
Musical work      & 30.15 & 49.99 & +19.84 \\
Natural place     & 11.54 & 30.77 & +19.23 \\
Plant             & 36.11 & 50.93 & +14.82 \\
Animal            & 25.00 & 38.89 & +13.89 \\
Artwork           & 22.54 & 32.57 & +10.03 \\
Fictional entity  & 34.16 & 42.97 & +8.81 \\
Person            & 22.40 & 30.71 & +8.31 \\
Book              & 24.94 & 33.00 & +8.06 \\
Food              & 46.35 & 52.73 & +6.38 \\
Landmark          & 27.78 & 33.78 & +6.00 \\
Movie             & 12.96 & 18.85 & +5.89 \\
Place of worship  & 31.86 & 37.13 & +5.27 \\
TV series         & 12.75 & 17.62 & +4.87 \\
Book series       &  8.82 & 10.58 & +1.76 \\
\midrule
\textsc{Macro Avg}& 24.81 & 34.32 & +9.51 \\
\bottomrule
\end{tabular}
\end{table}

\paragraph{Further Discussion}

Our finding that verifiable rewards activate dormant parametric knowledge and induce transferable reasoning strategies resonates with concurrent advances across several domains.
In the context of LLM reasoning, \citet{zhao2026know} show that metacognitive entropy calibration substantially improves verifiable RL reasoning by helping models better distinguish what they know from what they do not, aligning with our observation that explicit verification signals help surface latent knowledge.
Similarly, \citet{yu2026knowrlboostingllmreasoning} demonstrate that minimal-sufficient knowledge guidance can boost reasoning, complementing our entity-anchored approach where Wikidata aliases serve as the guiding rewards.
Beyond text, the RLVR paradigm shows analogous benefits in multimodal settings \citep{yuan2025segui,fakehr1,jiang2025ivy,jiang2025tabdsr}.
Together, these results suggest that the core mechanism underlying EA-RLVR, using verifiable rewards to induce reasoning rather than remembering knowledge mapping, may constitute a general principle applicable across modalities.
Looking forward, mechanistic interpretability methods such as the translation-switch discovery of \citet{wu2026finding} offer a promising lens for uncovering the internal circuits through which EA-RLVR activates parametric knowledge, a direction we leave for future work.


\section{Implementation Details}
\label{app:details}
\begin{table}[h!t!]
\centering
\caption{Languages used in our experiments, together with their ISO 639-1 codes, language--region locales, and English names.}
\label{tab:language-codes-single-column}
\resizebox{\linewidth}{!}{
\begin{tabular}{lll}
\toprule
\textbf{ISO 639-1} & \textbf{Locale} & \textbf{English Name} \\
\midrule
    ar & ar\_SA & Arabic (Saudi Arabia) \\
    de & de\_DE & German (Germany) \\
    es & es\_MX & Spanish (Mexico) \\
    fr & fr\_FR & French (France) \\
    it & it\_IT & Italian (Italy) \\
    ja & ja\_JP & Japanese (Japan) \\
    ko & ko\_KR & Korean (Korea) \\
    th & th\_TH & Thai (Thailand) \\
    tr & tr\_TR & Turkish (Turkey) \\
    zh & zh\_TW & Chinese (Taiwan, Traditional) \\
\bottomrule
\end{tabular}}
\end{table}

\paragraph{Data Construction.}
We conduct experiments on the XC-Translate benchmark across the ten language directions listed in Table~\ref{tab:language-codes-single-column}.
As the benchmark does not provide a dedicated training set, we repurpose the official validation set as our training split.
Table~\ref{tab:data-stats} summarizes the statistics of our data split.
Crucially, to strictly evaluate the model's ability to generalize rather than memorize, we ensure that \textbf{the training and test sets share no overlapping entities}.
The final setup comprises 7,278 examples for training and 49,606 examples for testing.

\begin{table}[h!t!]
\centering
\small
\caption{Statistics of the dataset used in our experiments. We utilize the official validation set of XC-Translate as our training split. The training and test sets are strictly disjoint in terms of entity coverage.}
\label{tab:data-stats}
\resizebox{0.9\linewidth}{!}{
\begin{tabular}{lrrr}
\toprule
\textbf{Language Pair} & \textbf{Train} & \textbf{Test} & \textbf{Total} \\
\midrule
English $\rightarrow$ Arabic & 722 & 4,546 & 5,268 \\
English $\rightarrow$ Chinese & 722 & 5,181 & 5,903 \\
English $\rightarrow$ French & 724 & 5,464 & 6,188 \\
English $\rightarrow$ German & 731 & 5,875 & 6,606 \\
English $\rightarrow$ Italian & 730 & 5,097 & 5,827 \\
English $\rightarrow$ Japanese & 723 & 5,107 & 5,830 \\
English $\rightarrow$ Korean & 745 & 5,081 & 5,826 \\
English $\rightarrow$ Spanish & 739 & 5,337 & 6,076 \\
English $\rightarrow$ Thai & 710 & 3,446 & 4,156 \\
English $\rightarrow$ Turkish & 732 & 4,472 & 5,204 \\
\midrule
\textbf{Total} & \textbf{7,278} & \textbf{49,606} & \textbf{56,884} \\
\bottomrule
\end{tabular}}
\end{table}

\paragraph{Training Implementation.}
We implement EA-RLVR using the \texttt{verl} library~\citep{verl}, a framework designed for efficient RLHF post-training. Unless otherwise specified, all RL experiments across different model scales (8B, 14B) and variants (Instruct, Thinking) use the unified set of hyperparameters reported in Table~\ref{tab:hyperparameters}.

\paragraph{Compute and Environment.}
All models were trained on 32 $\times$ NVIDIA H100 80GB GPUs. The 7k samples training for the 8B model takes approximately 12 hours, while the 14B model takes 24 hours. And the scaled training using full XC-Translate for the 8B model takes approximately 24 hours, while the 14B model takes 48 hours. We use \texttt{FlashAttention} for efficient computation~\citep{dao2022flashattn}.

\paragraph{SFT Baseline.}
For the SFT baseline, we fine-tune the base models on the same 7k training examples for 2 epochs, supervised by reference translation. We use a learning rate of 1e-6 with a cosine decay schedule and a global batch size of 64. We SFT our model using ms-swift framework~\citep{zhao2024swift}.

\paragraph{EA-RLVR Configuration.}
Our EA-RLVR optimization follows the critic-free policy gradient approach described in \S\ref{sec:method:rl}. 
We initialize the actor network with the Qwen3 weights post-trained by their official team, which endow the model basic reasoning capability.
During the rollout phase, we sample $G=16$ responses for each prompt to compute the group-normalized advantages.

Table~\ref{tab:hyperparameters} lists the detailed hyperparameters used for the RLVR stage.

\begin{table}[h]
    \centering
    \small
    \caption{Hyperparameters for EA-RLVR training.}
    \label{tab:hyperparameters}
    \begin{tabular}{lr}
        \toprule
        \textbf{Hyperparameter} & \textbf{Value} \\
        \midrule
        \multicolumn{2}{l}{\textit{Optimization}} \\
        Optimizer & AdamW \\
        Peak Learning Rate (Actor) & 1e-6 \\
        Learning Rate Scheduler & Cosine \\
        Warmup Ratio & 0.05 \\
        Weight Decay & 0.1 \\
        Train Batch Size & 512 \\
        PPO Mini Batch Size & 128 \\
        Total Training Steps & 500 \\
        \midrule
        \multicolumn{2}{l}{\textit{PPO / Policy Gradient}} \\
        Group Size ($G$) & 16 \\
        Clip Ratio ($\varepsilon_{low},\varepsilon_{high}$) & 3e-4,4e-4 \\
        Advantage Estimator & Group Normalization \\
        \midrule
        \multicolumn{2}{l}{\textit{Generation / Rollout}} \\
        Sampling Temperature & 1.0 \\
        Top-$p$ & 1.0 \\
        Max Sequence Length & 4096 \\
        \midrule
        \multicolumn{2}{l}{\textit{Reward Function}} \\
        Format Reward ($\alpha$) & 0.2 \\
        Length Tolerance ($\tau$) & 2.0 \\
        \bottomrule
    \end{tabular}
\end{table}

\paragraph{Prompt Format.}
We use the standard chat template of the Qwen3 family. For the input, we wrap the source sentence with the instruction: \texttt{``Translate the following sentence into \{tgt\_lang\}, provide only the translated text :...''}.
For the task-specific prompting ablation (Table~\ref{tab:prompt_ablation}), we replace the standard instruction with a three-step chain-of-thought prompt:
\texttt{``Translate the following text from \{src\_lang\} to \{tgt\_lang\}. First, identify any culturally specific entities (such as books, movies, places, or idioms). Second, deliberate on their conventional and culturally appropriate translations in \{tgt\_lang\}. Finally, provide the full translated text:...''}

\paragraph{Evaluation Configuration.}
We treat the reasoning process (i.e., ``thinking'') as an intrinsic capability of the models rather than a separate module. Consequently, we enable the thinking mode by default for all applicable models (e.g., Qwen3, Marco-o1 and GPT5-mini) and all settings including SFT and RAG. To ensure a fair comparison, we allocate a unified maximum generation budget of 4,096 tokens for all experiments. Regarding decoding strategies, we follow the best practices established by DeepSeek-R1~\citep{deepseekai2026deepseekr1,2025qwen3}, setting the sampling temperature to 0.6 and top-$p$ to 0.95. This specific configuration is critical, as lower temperatures (e.g., greedy decoding) tend to induce severe repetition loops and infinite generation behaviors in reasoning-heavy models. Finally, to ensure statistical reliability, all reported results for open-weight models are averaged over three independent runs ($pass@1$).

\section{Unbiased $pass@k$ Estimator}
\label{app:passk_estimator}

While $pass@k$ is defined as the probability of generating at least one correct sample in $k$ attempts, directly computing this probability typically requires a very large number of samples to reduce variance. 
To evaluate $pass@k$ efficiently, we follow the method proposed by \citet{chen2021codex}. instead of just sampling $k$ times, we generate a larger number of samples $n$ (where $n \ge k$) for each input and count the number of correct samples $c$. The unbiased estimator for $pass@k$ is then calculated as:

\begin{equation}
\label{eq:passk_estimator}
\text{pass}@k := \mathbb{E}_{\text{problems}}\left[1 - \frac{\binom{n-c}{k}}{\binom{n}{k}}\right]
\end{equation}

where $\binom{n}{k}$ denotes the number of combinations of choosing $k$ items from a set of $n$. Mathematically, this formula calculates the probability that a randomly chosen subset of size $k$ contains at least one correct answer, derived from the complement of the probability that all $k$ chosen samples are incorrect (i.e., chosen from the $n-c$ incorrect samples).

In our experiments, we set the total sample budget $n=128$ and evaluate $pass@k$ for $k \in \{1, \dots, 128\}$. If $n-c < k$, the estimator returns $1.0$, as it is impossible to choose $k$ incorrect samples.

\begin{figure*}[bh!]
    \centering
    \includegraphics[width=1\linewidth]{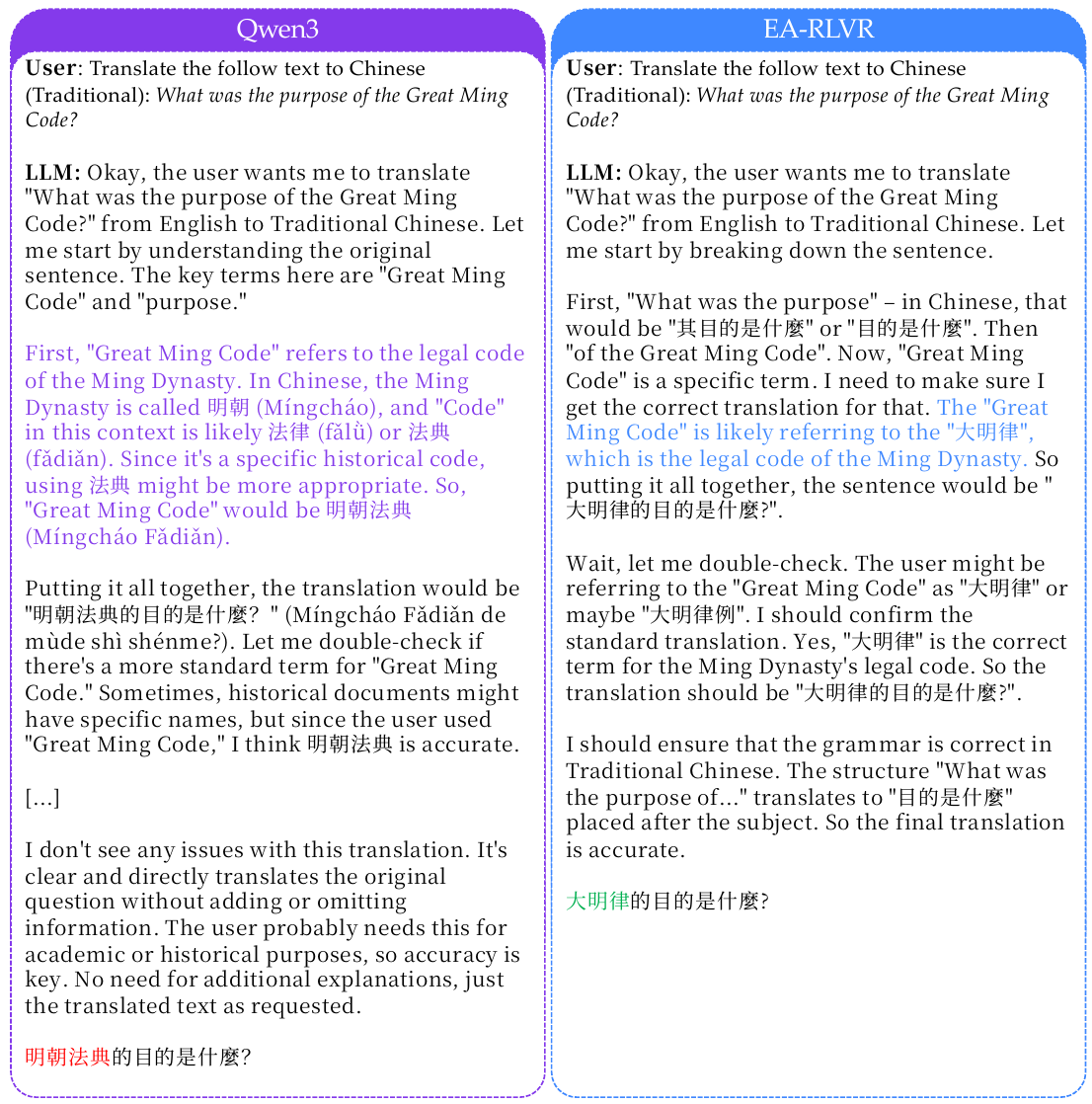}
    \caption{Case study 1: Canonicalization of Historical Terminology.}
    \label{fig:case1}
\end{figure*}

\begin{figure*}[bh!]
    \centering
    \includegraphics[width=1\linewidth]{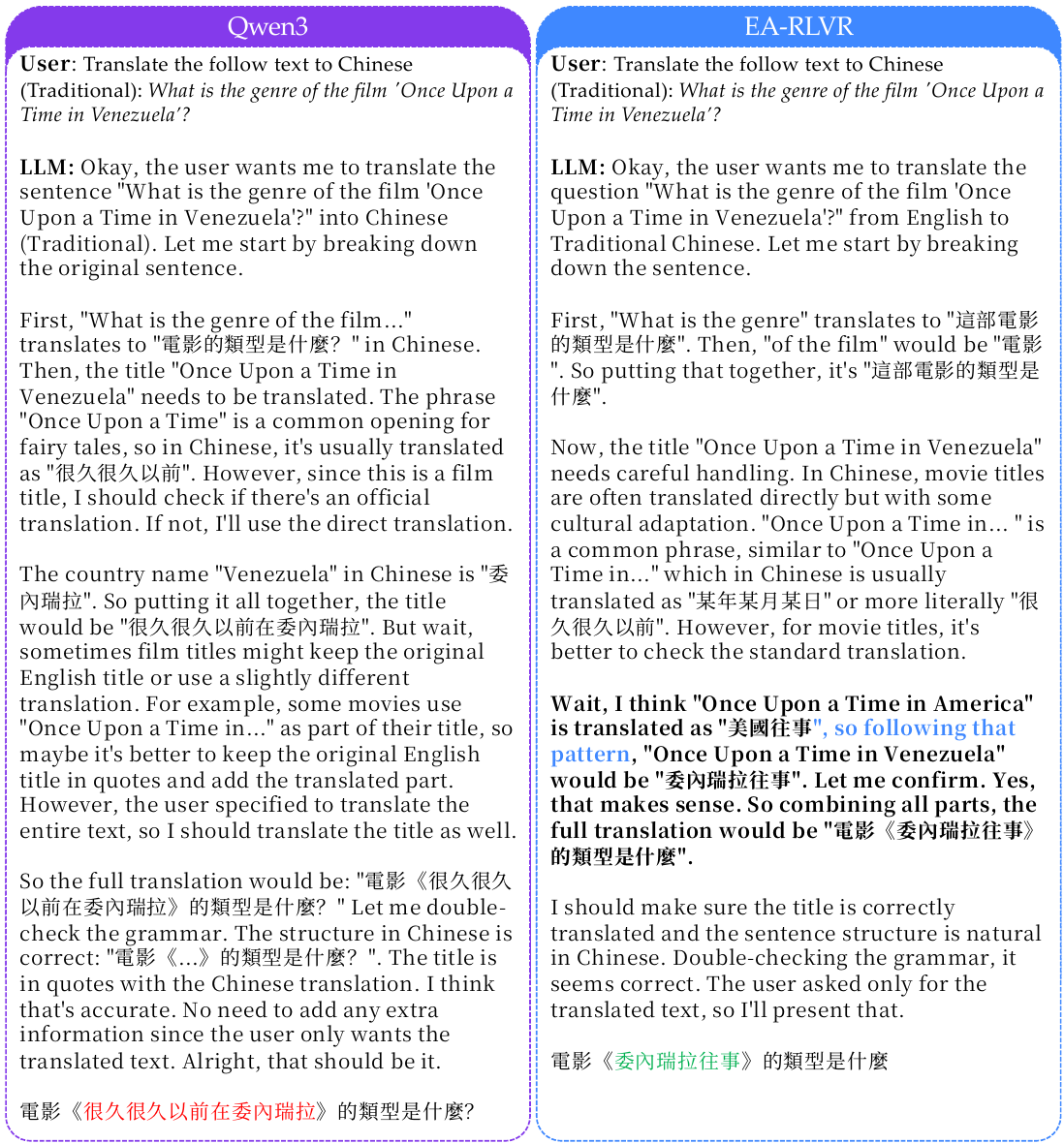}
    \caption{Case study 2: Analogical Reasoning for Cultural Conventions.}
    \label{fig:case2}
\end{figure*}

\begin{CJK*}{UTF8}{gbsn}
\section{Case Study}
\label{sec:case_study}
\paragraph{Qualitative Analysis}
We present a qualitative analysis of four representative cases to illuminate the mechanism by which EA-RLVR improves entity translation. By examining the generated reasoning traces (denoted as \texttt{LLM}), we identify a distinct shift in cognitive patterns: while the baseline model (Qwen3-8B) relies on \textit{literal semantic composition}, EA-RLVR exhibits \textit{entity-aware deliberation} and \textit{domain-specific retrieval}.

\paragraph{Canonicalization of Historical Terminology (Case 1).}
In Case 1, the user asks for the translation of ``Great Ming Code''.
The baseline Qwen3 adopts a compositional approach, translating ``Great Ming'' ($\rightarrow$ 明朝) and ``Code'' ($\rightarrow$ 法典) separately, resulting in the descriptive but non-standard phrase ``明朝法典'' (Ming Dynasty Code).
In contrast, EA-RLVR's reasoning trace explicitly triggers a hypothesis check: \textit{``Great Ming Code is likely referring to the 大明律... I should confirm the standard translation.''}.
By treating the phrase as a rigid proper noun rather than a translatable sentence fragment, EA-RLVR successfully retrieves the historiographically correct term ``大明律''.

\paragraph{Analogical Reasoning for Cultural Conventions (Case 2).}
Case 2 illustrates how EA-RLVR leverages parametric knowledge for style transfer.
For the film title ``Once Upon a Time in Venezuela'', Qwen3 defaults to a dictionary translation of the idiom ``Once Upon a Time'' ($\rightarrow$ 很久很久以前), missing the cinematic context.
EA-RLVR, however, employs \textbf{analogical reasoning}. The thinking trace reveals a crucial intermediate step: it recalls a prototype entity, \textit{``Once Upon a Time in America is translated as 美國往事''} and applies this naming convention to the target entity, synthesizing the culturally attuned title ``委內瑞拉往事'' (Venezuela Chronicles/Past). This demonstrates the model's ability to map new entities to existing cultural schemas.

\paragraph{Domain-Specific Disambiguation (Case 3 \& 4).}
Polysemy poses a major challenge in entity translation. In Case 3, Qwen3 fails to resolve the term ``Mahavira Hall'' within a Buddhist context. Confused by the association of ``Mahavira'' with Jainism, it resorts to a phonetic transliteration ``馬哈維拉廳''.
EA-RLVR correctly identifies the domain constraints: \textit{``In Chinese Buddhist terminology... Mahavira is often transliterated as 大雄''}. It successfully navigates the semantic shift of ``Mahavira'' (Great Hero) in Buddhism to produce the correct temple hall name ``大雄寶殿''.
Similarly, in Case 4 (French), EA-RLVR shows a willingness to localize media titles (``Emi magique'') rather than leaving them in English, reflecting a deeper adherence to target-language publication norms.

\paragraph{Summary of Learned Reasoning Pattern.}
Across these cases, a consistent meta-strategy emerges.
The standard SFT/Base model tends to optimize for $P(y|x)$ at the token level, favoring high-frequency phrases (fluency) over factual precision.
EA-RLVR, driven by the entity-anchored reward, learns a \textbf{``Pause-Retrieve-Verify''} loop. It (1) detects potential cultural entities, (2) suspends immediate translation to search its parametric memory for domain equivalents, and (3) verifies the candidate against the target cultural context. This shift from \textit{translating meaning} to \textit{matching entities} is the core driver of the observed performance gains.
\end{CJK*}

\begin{figure*}
    \centering
    \includegraphics[width=1\linewidth]{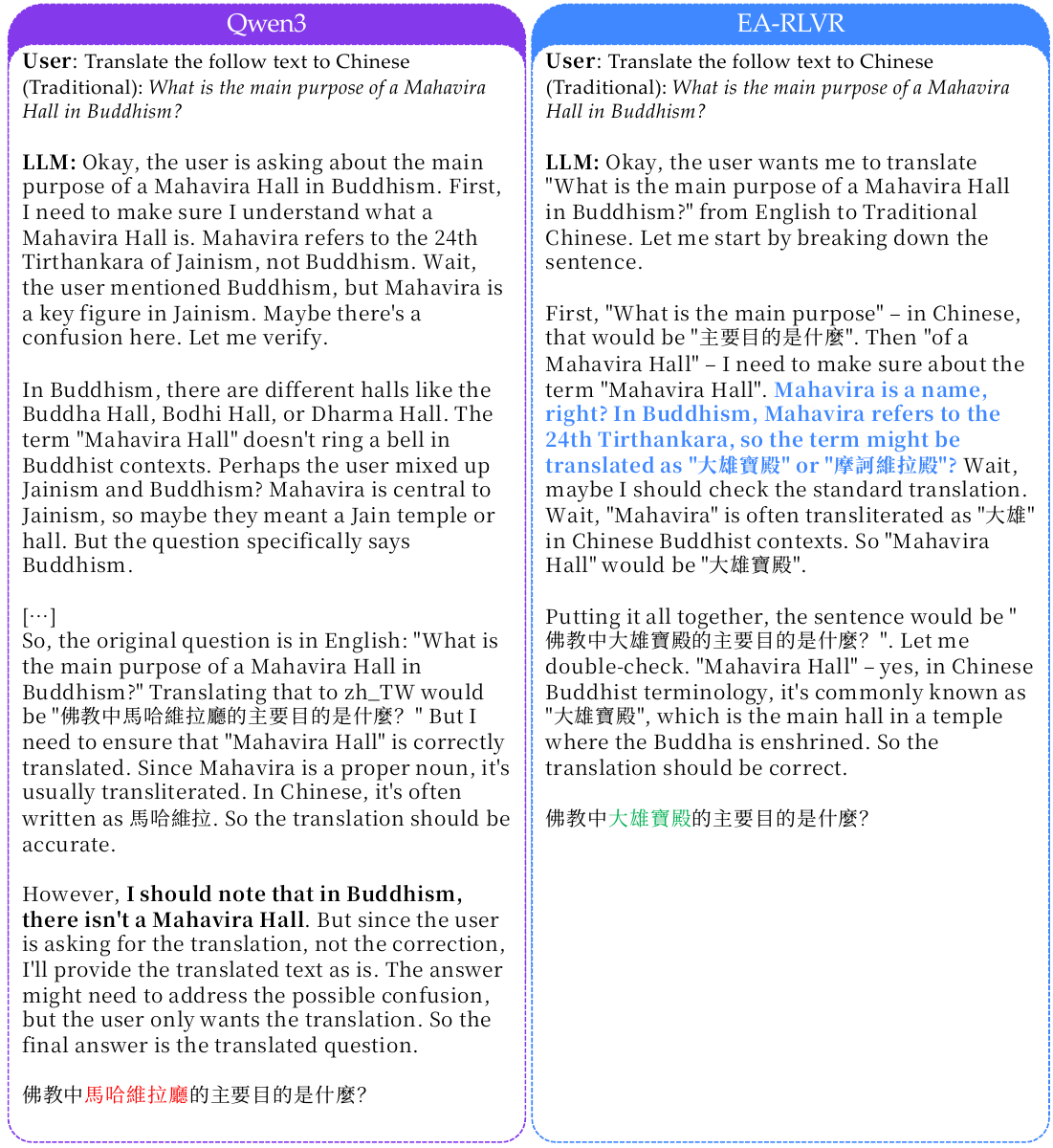}
    \caption{Case study 3: Domain-Specific Disambiguation .}
    \label{fig:case3}
\end{figure*}
\begin{figure*}
    \centering
    \includegraphics[width=1\linewidth]{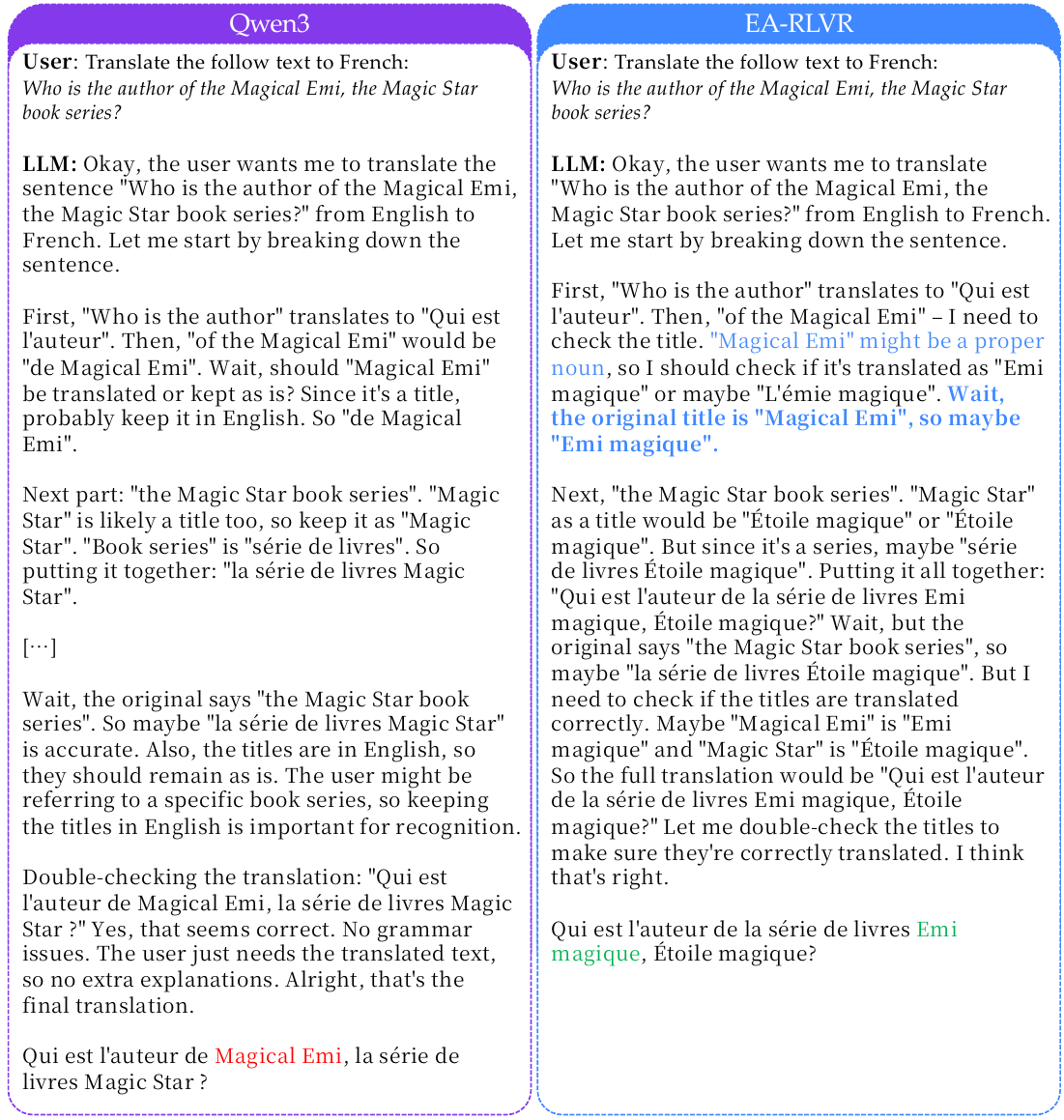}
    \caption{Case study 4: Domain-Specific Disambiguation .}
    \label{fig:case4}
\end{figure*}


\end{document}